\documentclass{article} % For LaTeX2e
\usepackage{iclr2027_conference,times}

\usepackage{amsmath,amsfonts,bm}

\def\eqref#1{equation~\ref{#1}}
\def\1{\bm{1}}

\DeclareMathAlphabet{\mathsfit}{\encodingdefault}{\sfdefault}{m}{sl}
\SetMathAlphabet{\mathsfit}{bold}{\encodingdefault}{\sfdefault}{bx}{n}

\usepackage{hyperref}
\usepackage{url}

\usepackage{algorithm}
\usepackage{algorithmic}
\usepackage{subfiles}
\usepackage{booktabs}       % professional-quality tables
\usepackage{amsmath,mathtools}
\usepackage{amssymb}
\usepackage{amsfonts}
\usepackage{multirow}
\usepackage{multicol}
\usepackage{xcolor}
\usepackage{adjustbox}
\usepackage{amsthm}
\theoremstyle{definition}

\theoremstyle{remark}

\usepackage{pifont}
\usepackage{caption}
\usepackage{subcaption}
\usepackage{wrapfig}
\usepackage{simplebnf}
\usepackage{tcolorbox}
\usepackage{tikz}
\usepackage{tikz-qtree}
\usepackage{dashrule}
\tcbuselibrary{listings,breakable}
\tcbset{listing engine=listings,colframe=black,colback=white,size=small}
\usepackage{enumitem}

\usepackage{booktabs,multirow,tabularx,makecell,array}
\newcolumntype{Y}{>{\raggedright\arraybackslash}X} % wrapping text column

\usepackage[table]{xcolor}

\usepackage{graphicx}
\usepackage{svg} % only needed if using .svg files
\usepackage{twemojis}

\definecolor{LightYellow}{RGB}{255,250,208}
\definecolor{LightGreen}{RGB}{215, 245, 203}
\definecolor{LightBlue}{RGB}{208, 245, 243}
\definecolor{LightPurple}{RGB}{224,223,255}
\definecolor{LightGrey}{RGB}{225,225,225}

\usepackage{soul}
\usepackage{needspace}
\tcbuselibrary{skins}
\definecolor{qexprompt}{HTML}{888780}
\definecolor{qexours}{HTML}{185FA5}
\definecolor{qexrar}{HTML}{5F5E5A}
\newcommand{\qexhl}[1]{{\sethlcolor{LightGreen}\hl{#1}}}
\newcommand{\qexhead}[1]{\par\smallskip\noindent\textbf{#1}\par\nopagebreak}
\newcommand{\qexitem}[1]{\par\noindent\hangindent=1.5em\hangafter=1\makebox[1.5em][l]{#1}\ignorespaces}
\newcommand{\qexsubitem}[1]{\par\noindent\hangindent=3em\hangafter=1\hspace*{1.5em}\makebox[1.5em][l]{#1}\ignorespaces}
\newtcolorbox{qexbox}[2]{enhanced, breakable, beforeafter skip balanced=0.5\baselineskip,
  colframe=#1, colback=white,
  colbacktitle=#1!10!white, coltitle=#1!70!black, boxrule=0.5pt, arc=1pt,
  left=6pt, right=6pt, top=4pt, bottom=4pt, toptitle=2pt, bottomtitle=2pt,
  fontupper=\scriptsize, fonttitle=\small, title={\textbf{#2}},
  before upper={\setlength{\parskip}{1pt}\setlength{\parindent}{0pt}}}
\newcommand{\qexeq}[1]{\par{\centering\adjustbox{max width=\linewidth}{$\displaystyle #1$}\par}}
\newcommand{\qexeqhl}[1]{\par{\centering\adjustbox{max width=\linewidth}{\setlength{\fboxsep}{1pt}\colorbox{LightGreen}{$\displaystyle #1$}}\par}}

\title{\raisebox{-0.25cm}{\includegraphics[width=1.0cm, height=1.0cm]{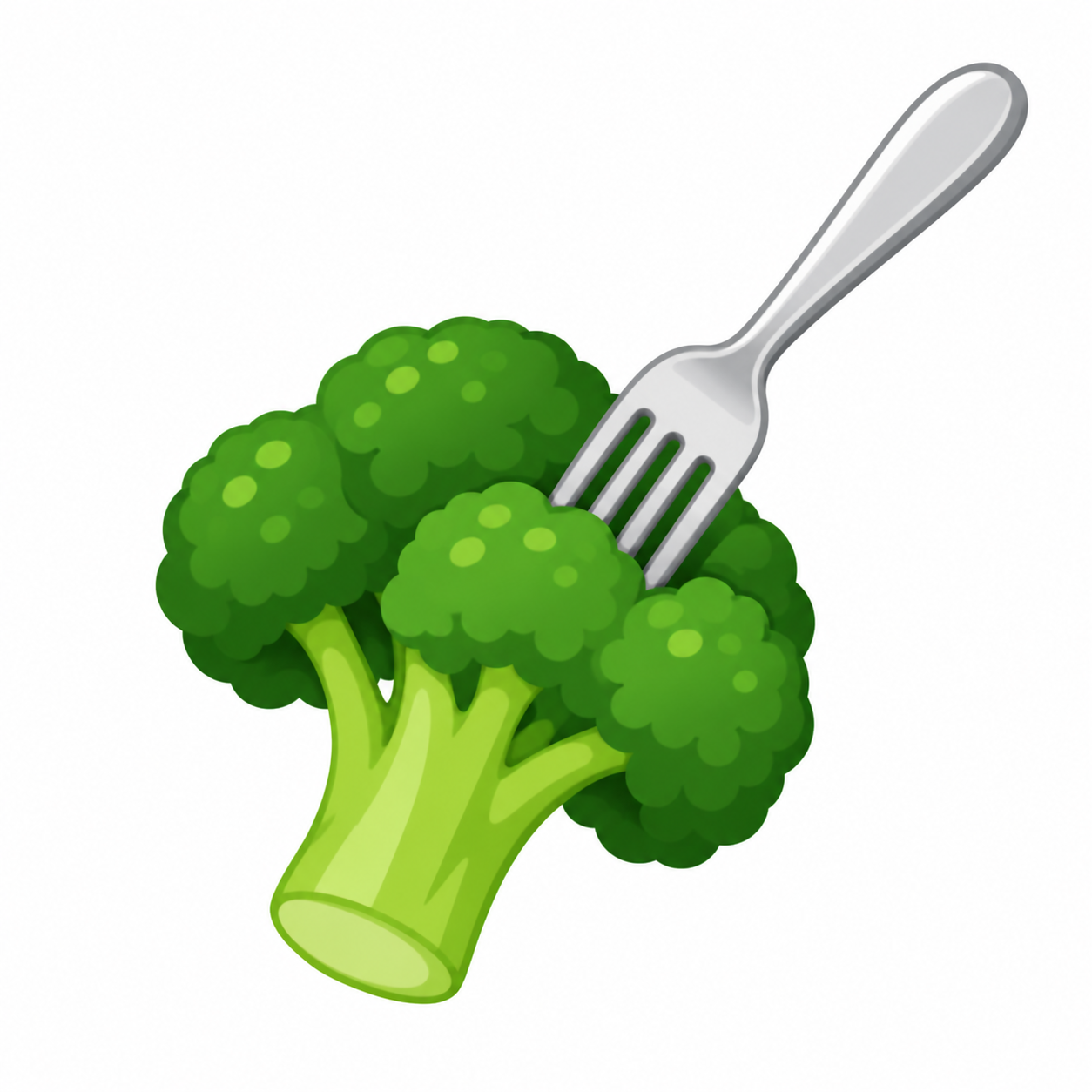}}Eat Your Precision: Precision Objective \\ Mitigates Reward Hacking in Rubric-based RL}
\author{
Hyun Ryu$^{*,1,2}$~~Jongheon Kim$^{*,1}$~~Yiqing Xie$^{2}$~~Seungone Kim$^{2}$~~Carolyn Ros\'e$^{\dagger,2}$~~Eunho Yang$^{\dagger,1}$ \\
$^{1}$KAIST AI~~~$^{2}$Carnegie Mellon University Language Technologies Institute\\
\texttt{\{ryuhyun1905,jongheon,eunhoy\}@kaist.ac.kr}\\
\texttt{\{hyunr,yiqingxi,seungonk,cprose\}@cs.cmu.edu}
}

\iclrfinalcopy % Uncomment for camera-ready version, but NOT for submission.
\begin{document}

\maketitle
% Author contribution/advising note at the bottom of page 1 (no marker in the text)
\begingroup
\renewcommand{\thefootnote}{}
\footnotetext{\textsuperscript{*}Equal Contribution \textsuperscript{\textdagger}Equal advising}
\addtocounter{footnote}{0}
\endgroup

\begin{abstract}
% \vspace{-2mm}
Rubric-based rewards enable effective reinforcement learning on non-verifiable tasks. However, because these rewards are typically defined as a weighted measure of rubric coverage, models are trained primarily to maximize a \textit{recall} objective. As a result, RL-trained models produce responses that are more complete but also more verbose and contain more unnecessary or factually incorrect content, leading them to be less preferred than responses from the base model.
To address this issue, we present the \textsc{Fork} reward, which jointly optimizes recall and a novel precision objective, which requires each unit of a model response to be both necessary and factually correct, as measured by rubric-driven necessity credits and a factual-correctness grader, respectively.
Our experimental results show that \textsc{Fork} outperforms existing rubric-based RL methods and achieves Pareto-optimal performance that balances rubric coverage and overall response quality.
Compared to the original recall-based reward, \textsc{Fork} substantially improves precision and reduces response length while achieving similar rubric coverage, which leads to significantly higher win rate compared to the base policy (e.g., from 10.6\% to 68.0\% on RubricHub-Medical and from 22.1\% to 74.1\% on HealthBench).
\footnote{The source code is available at \url{https://github.com/JHKim631/FORK}.}
% \vspace{-3mm}
\end{abstract}

\section{Introduction}
\label{sec:intro}
% rar 소개
% 리워드 해킹
% 우리의 해결 방안 -- 방법론 설명
Reinforcement learning (RL) with verifiable rewards has led to rapid advances in AI for math and coding, where the correctness of the answers can be reliably measured and used as reward signals.
In comparison, for non-verifiable tasks such as open-ended generation, the evaluation criteria for model responses can be diverse and sometimes subjective, which makes it difficult to define rewards in RL.
To resolve this issue, \citet{rar} and \citet{checklists} leverage rubric-based RL, which first generates a set of rubrics that model responses should satisfy and then uses the weighted coverage of these pre-defined rubrics as rewards.
% Here, the ... % 루브릭은 대부분 presence-based이다. (루브릭 성격 설명) -> 메소드에 언급됨.

\begin{figure}[t]
  \vspace{-5mm}
  \centering
  \includegraphics[width=\textwidth]{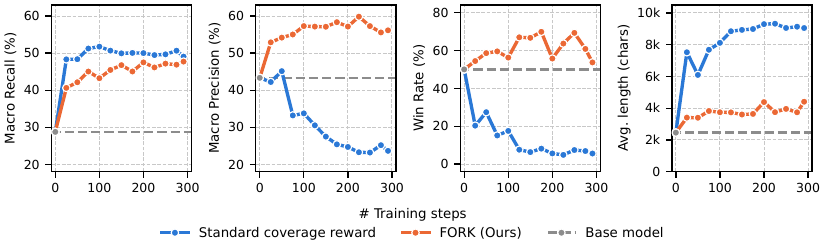}
  \vspace{-7mm}
  \caption{
  \textbf{Reward hacking in rubric-based RL and our mitigation method.}
  With the standard coverage reward, recall (i.e., weighted coverage of rubrics) quickly saturates, while precision and win rate against a base model continually decreases and responses become more verbose.
  With a precision objective included in the reward (\textsc{Fork}), the model reaches similar recall while gradually increasing precision and win rate and generating less verbose responses.
  Here, we train Qwen2.5-7B-Instruct on RubricHub-Medical and evaluate it on an in-domain test set.
  % using Qwen3.6-35B-A3B as an LLM grader
  }
  \label{fig:intro}
  \vspace{-5mm}
\end{figure}

However, such rubric-coverage-based reward is easily hacked.
As shown in Figure~\ref{fig:intro}, when a model is only optimized towards maximizing the coverage of pre-defined rubrics, 
we observe that although the model responses become more complete, their factual correctness and necessity decrease, and the responses become more verbose. This makes them less preferred than base policy responses~\citep{hacking}.
% Specifically, compared to an initial base model responses, completeness of rubric-based RL-trained model responses increases following the current training objective, but factual correctness and necessity of model responses decrease, and model responses are also getting more verbose~\citep{hacking}.
% 해킹 논문에서 보고한 세가지 현상 모두 쓰기? A. 굳이?
This reward hacking behavior is not surprising since rubrics as rewards currently behave like a \textit{recall} objective. The missing part is a \textit{precision} objective: whether the content of a model response is factually correct and necessary.
% We empirically verify that during rubric-based RL training, recall (i.e., the rubric coverage score) increases whereas precision continually decreases.
% 여기에 fig.1 -- precision,recall vs. # training steps -- rar은 precision이 점점 감소함. 우리 것은 precision도 점점 증가함. 다른 baseline (universal rubrics)은 우리 것보다는 비효율적.
% % precision을 정의하는 다른 차원은 없나? 3가지를 너무 당연하게 사용함. - 이 세가지에 대한 justify가 필요함
% % -> 이거는 리워드 해킹 논문에서 차용하면 됨.

% Instead of adding more rubrics or modifying existing rubrics~\citep{onlinerubrics, drtulu, rubricdropout}, 
With this motivation,
we propose \textsc{Fork}, which converts rubric-based rewards from a recall score into an $F_\beta$ score that aggregates recall and precision via their weighted harmonic mean.
% 방법론 자세히 설명
Given a set of rubrics, 
% we describe how to measure the precision of free-form responses with an LLM grader.
to measure the precision,
we first divide a model response into a sequence of units and then determine the necessity and factual correctness of each unit by prompting an LLM grader. 
% However, since we observe that necessity judgments are inaccurate
To improve the accuracy of necessity judgments, which we observe as a bottleneck\footnote{Necessity judgments of all candidate training graders show only moderate correlation with stronger reference graders, whereas factual correctness judgments show stronger correlation (Table~\ref{tab:grader-selection}).}, we develop a credit-based method in which an LLM grader matches rubrics to units, assigns each unit a necessity credit, and determines unit necessity based on a dynamic threshold value.

% 실험결과 설명 (이걸 충분히 쓰자)
We demonstrate the effectiveness of \textsc{Fork} on both in-domain datasets (the test splits of RubricHub-Medical and RubricHub-Science, \citet{rubrichub}) and external benchmarks (HealthBench, \citet{healthbench} and ResearchQA, \citet{researchqa}).
% Experiments show that on an in-domain RubricHub~\citep{rubrichub} test set,  
Compared to the standard coverage reward, \textsc{Fork} significantly improves the factual correctness 
and necessity of responses (e.g., reducing the response length by a factor of 2.6) while maintaining similar rubric coverage.
% It also reduces response length by a factor of 2.6 compared to the standard coverage reward. 
These improvements also make responses from rubric-based RL-trained models more preferred than base policy responses.
% Our method also effectively generalizes to external benchmarks, HealthBench~\citep{healthbench} and ResearchQA~\citep{researchqa}.

Further analysis supports the effectiveness of \textsc{Fork}.
While rubric-coverage rewards~\citep{rar,onlinerubrics,checklists} have a negative effect on satisfaction of absence-based rubrics, our method improves satisfaction of both presence-based and absence-based rubrics, enabling models to add necessary content and suppress irrelevant content at the same time.
We also verify that a model trained with our method remains preferred to the base model under different judge prompts and remains more effective than other rubric-based RL methods when evaluated with different LLM graders.
% Furthermore, when aggregating precision and recall into a single reward, our $F_\beta$ score outperforms simple averaging.

% 핵심 기여도 요약
Our key contributions are:
\begin{enumerate}[leftmargin=7mm]
\vspace{-2mm}
    \item We propose \textsc{Fork}, which mitigates reward hacking in rubric-based RL training by incorporating a precision objective into a standard coverage objective.
    \item We develop a reliable method for using an LLM grader to measure the precision of free-form responses given a set of rubrics to be satisfied.
    \item 
    Experiments show that on multiple public benchmarks, compared to the standard coverage reward, \textsc{Fork} significantly improves precision while maintaining similar rubric coverage, 
    which leads to substantial improvements in the preference score (i.e., win rate compared to the base policy).
    % \item 
    \textsc{Fork} also outperforms other rubric-based RL methods, achieving Pareto-optimal performance that balances rubric coverage and overall response quality.
\vspace{-2mm}
\end{enumerate}

\section{Related Works}
\label{sec:related}

\paragraph{Rubric-Based Reinforcement Learning.}
Rubric-based methods decompose response quality into explicit, prompt-specific criteria, providing structured supervision for tasks without readily verifiable outcomes~\citep{rar,checklists}.
Rubrics as Rewards uses expert-guided, instance-specific rubrics as on-policy RL rewards, exploring both explicit aggregation of criterion scores and implicit aggregation by an LLM grader~\citep{rar}.
Reinforcement Learning from Checklist Feedback (RLCF) evaluates instruction-specific requirements using LLM graders and specialized verification programs, and converts aggregated checklist scores into preference pairs for direct preference optimization~\citep{checklists}.
RubricHub scales rubric supervision through coarse-to-fine generation that combines principle-guided synthesis, multi-model aggregation, and difficulty evolution, supporting rubric-based RL~\citep{rubrichub}.
These methods establish rubrics as a practical interface for specifying training objectives.
We build on this paradigm while distinguishing coverage of specified requirements from the quality of all generated content.

\paragraph{Rubric Specification and Reward Hacking.}
Rubric-based optimization remains vulnerable to incomplete reward specifications.
\citet{hacking} distinguish verifier failure from rubric-design limitations: even with strong verification, rubric-based judges can prefer RL-trained responses that rubric-free judges rate worse overall, with improved completeness but degraded factual correctness and necessity.
Several approaches address insufficiently specified requirements.
RLCF adds universal criteria against excessive or off-topic content after observing verbose preambles under checklist optimization~\citep{checklists}.
\citet{chasing} refine rubrics by contrasting high-quality, diverse responses, improving discrimination in the high-reward region and delaying over-optimization.
OnlineRubrics augments existing criteria through comparisons between current-policy and control-policy responses~\citep{onlinerubrics}.
DR Tulu maintains search-grounded, evolving rubrics that capture both newly discovered useful content and undesirable behaviors observed in rollouts~\citep{drtulu}.
AMARIS further uses persistent training diagnostics to revise criteria and weights in response to recurring failures~\citep{amaris}.
These specification-refinement strategies motivate a complementary question: how should generated content be evaluated when coverage requirements remain incomplete?
Our method supplements rubric coverage with unit-level assessments of necessity and factual correctness, rather than relying on criterion expansion alone, and does not require online rubric updates.

\paragraph{Fine-Grained Evaluation and Precision--Recall Rewards.}
Content-level evaluation provides a complementary perspective on response quality.
\citet{factscore} measure factual precision as the proportion of generated atomic facts supported by a reliable knowledge source.
\citet{finesure} combine sentence-level faithfulness assessment with keyfact--sentence alignment to measure completeness and conciseness, distinguishing coverage of target information from the fraction of generated sentences contributing that information.
Related work directly incorporates precision--recall signals into policy optimization.
\citet{clinical} train clinical note generators with a claim-level $F_1$ reward combining entailment-based precision against the source dialogue and recall over dialogue-derived claims.
For video captioning, \citet{cure} verify atomic claims against video evidence and balance factual precision with reference-claim coverage, assigning attenuated credit to supported but unmatched details and bounding their contribution to recall.
Our work adapts this content-level perspective to rubric-based RL: necessity is derived from a unit's contribution to prompt-specific coverage requirements, and precision counts units that satisfy both necessity and factual correctness.
Combining this precision with rubric-coverage recall through an $F_\beta$ reward couples what a response should cover with whether its generated content is warranted and correct.

\section{Fork: Precision-Aware Rubric Reward}
\label{sec:method}

\begin{figure}[t]
  \vspace{-5mm}
  \centering
  \includegraphics[width=1.0\textwidth]{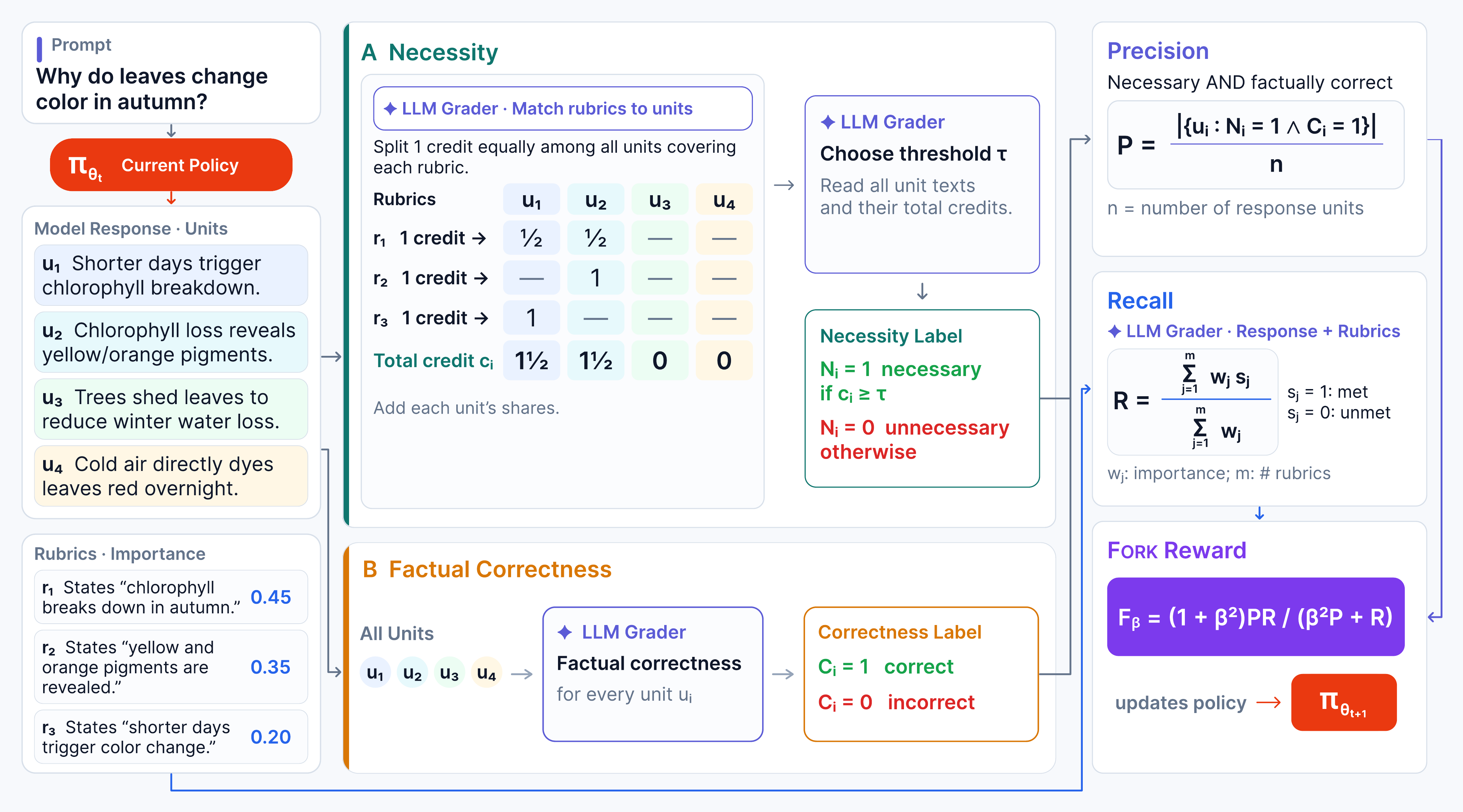}
  \caption{
  \textbf{Training with \textsc{Fork}.}
  Given a prompt, a policy model samples a response, which is then divided into units.
  An LLM grader matches rubrics to units to assign each unit a necessity credit and determines a credit threshold for unit necessity.
  In parallel, the grader assesses the factual correctness of all units.
  Precision is defined as the fraction of necessary and factually correct units, and recall follows the standard definition of weighted rubric coverage.
  We update the policy model using their weighted harmonic mean as a reward.
  }
  \label{fig:overview}
  \vspace{-5mm}
\end{figure}

In this section, we propose \textsc{Fork}, which can mitigate reward hacking in rubric-based RL by incorporating a precision objective (Figure~\ref{fig:overview}).
We first provide background on the standard rubric-coverage reward formulation (Section~\ref{subsec:coverage-reward}), describe how to measure the precision of model responses (Section~\ref{subsec:response-precision}), and introduce \textsc{Fork} as a new reward formulation for rubric-based RL (Section~\ref{subsec:rubric-fb}).

\subsection{Preliminary: Rubric-Coverage Reward}
\label{subsec:coverage-reward}

Given a prompt \(x\), the current policy \(\pi_{\theta_t}\) generates a free-form response $y \sim \pi_{\theta_t}(\cdot \mid x)$.
Let
\[
\mathcal{R}(x) = \{(r_j,w_j)\}_{j=1}^{M}
\]
denote the coverage rubrics associated with \(x\), where \(r_j\) specifies a desired element of the response and \(w_j>0\) denotes its importance. For each rubric $r_j$, an LLM grader produces a binary coverage label
\[
s_j(x,y)
=
\mathbb{I}\!\left[
y \text{ satisfies } r_j
\right]
\in\{0,1\}.
\]
The conventional rubric-coverage reward is the importance-weighted fraction of satisfied rubrics:
\begin{equation}
\operatorname{Rec}(x,y)
=
\frac{\sum_{j=1}^{M} w_j s_j(x,y)}
     {\sum_{j=1}^{M} w_j}.
\label{eq:rubric-recall}
\end{equation}
We refer to this quantity as \textit{rubric recall} (or simply recall), since it measures how much of the desired content is covered by the response.
Rubric recall, however, evaluates mainly whether the requested content is present\footnote{RubricHub~\citep{rubrichub}, a widely adopted dataset for rubric-based RL training, consists of 90.2\% presence-based rubrics and 8.6\% absence-based rubrics~\citep{hacking}.}. It does not penalize content that is redundant, unnecessary, or factually incorrect unless specific failure cases are covered by rubrics. Consequently, a response $y$ can obtain a high coverage reward while containing substantial low-quality content. We therefore complement rubric recall with a unit-level notion of precision.

\subsection{Measuring Precision of Free-Form Responses}
\label{subsec:response-precision}

\paragraph{Response Unitization.}
We first transform the response into an ordered sequence of units:
\[
\mathcal{U}(y)
=
(u_1,\ldots,u_n).
\]
For simplicity, we define a unit as approximately a single sentence in the original response\footnote{We also divide the response into units by line breaks (`\textbackslash n') and report the results in Section~\ref{subsec:analysis}.}. We preserve the original wording and substantive content, making only minimal substitutions needed to make each unit self-contained. For example, a pronoun may be replaced with its antecedent while retaining the remainder of the original sentence. Purely functional or organizational material---such as greetings, section headers, discourse markers (e.g., ``first'' or ``in summary''), and transition-only sentences or phrases---is discarded. Unitization therefore isolates the substantive content of the response without summarizing or rewriting it.
We automate this process using an LLM unitizer\footnote{Because response unitization is a trivial task, we use the same model for response unitization and grading for simplicity.}.

\paragraph{Credit-based Necessity.}
We determine whether each unit contributes necessary content toward the coverage rubrics. For every rubric--unit pair $(r_j, u_i)$, an LLM grader predicts
\[
m_{ji}
=
\mathbb{I}\!\left[
u_i \text{ contributes to addressing } r_j
\right]
\in\{0,1\}.
\]
Let $K_j=\sum_{i=1}^{n}m_{ji}$ be the number of units matched to rubric \(r_j\). We assign each rubric a total unit-credit budget of one and divide this credit equally among its matched units. The credit assigned by rubric \(r_j\) to unit \(u_i\) is
\begin{equation}
a_{ji}
=
\begin{cases}
\dfrac{m_{ji}}{K_j}, & K_j>0,\\[4pt]
0, & K_j=0.
\end{cases}
\label{eq:unit-rubric-credit}
\end{equation}
The accumulated necessity credit of \(u_i\) is then $c_i = \sum_{j=1}^{M}a_{ji}$.
This allocation discounts redundant units: if \(K_j\) units address the same rubric, each receives only \(1/K_j\) credit from that rubric. At the same time, a unit may accumulate credit from multiple rubrics when it contributes to several desired elements.
Given the accumulated necessity credits for all units, an LLM grader determines a threshold \(\tau>0\), which defines the binary necessity label
\begin{equation}
N_i
=
\mathbb{I}[c_i\geq\tau].
\label{eq:necessity-label}
\end{equation}
Thus, \(N_i=1\) indicates that \(u_i\) contributes sufficient rubric-grounded content, whereas \(N_i=0\) indicates that its contribution is insufficient or redundant.

\paragraph{Factual Correctness.}
Independently of necessity, an LLM grader assigns each unit a binary factual correctness label
\begin{equation}
C_i
=
\mathbb{I}\!\left[
u_i \text{ is factually correct given } x
\right]
\in\{0,1\}.
\label{eq:correctness-label}
\end{equation}
We use a strict unit-level criterion: \(C_i=1\) only when the factual claims expressed by \(u_i\) are correct; a unit containing a factual error receives \(C_i=0\).

\paragraph{Response Precision.}
A unit contributes positively to precision only if it is both necessary and factually correct. We therefore define the precision of the response as
\begin{equation}
\operatorname{Prec}(x,y)
=
\frac{1}{n}
\sum_{i=1}^{n}N_iC_i.
\label{eq:response-precision}
\end{equation}

\subsection{Fork Reward}
\label{subsec:rubric-fb}

We combine response precision and rubric recall using their weighted harmonic mean (i.e., $F_\beta$ score):
\begin{equation}
r_{\mathrm{FORK}}(x,y)
=
\begin{cases}
\dfrac{
(1 + \beta^2)\,\operatorname{Prec}(x,y)\operatorname{Rec}(x,y)
}{
\beta^2 \operatorname{Prec}(x,y)+\operatorname{Rec}(x,y)
},
&
\operatorname{Prec}(x,y)+\operatorname{Rec}(x,y)>0,
\\[10pt]
0, & \text{otherwise}.
\end{cases}
\label{eq:rubric-fb}
\end{equation}
The harmonic mean favors responses that jointly cover the desired rubrics and express that content using necessary, factually correct units. In particular, increasing coverage cannot compensate for arbitrarily low precision, and vice versa.
Here, $\beta>1$ gives more weight to recall and $\beta<1$ gives more weight to precision.
The resulting score serves as the reward for updating the policy. The corresponding learning objective is
\begin{equation}
\max_{\theta}\;
\mathbb{E}_{x\sim\mathcal{D},\,
            y\sim\pi_{\theta}(\cdot\mid x)}
\left[
r_{\mathrm{FORK}}(x,y)
\right],
\label{eq:rubric-fb-objective}
\end{equation}
producing an updated policy \(\pi_{\theta_{t+1}}\) from the current policy \(\pi_{\theta_t}\).

\section{Experiments}
\label{sec:exp}

Our experiments address four research questions:
\textbf{(RQ1)} Can \textsc{Fork} improve response precision and overall quality while maintaining comparable rubric coverage?
\textbf{(RQ2)} Do these improvements generalize to external benchmarks?
\textbf{(RQ3)} How do precision estimation and precision--recall aggregation affect performance?
\textbf{(RQ4)} Are the observed improvements robust to changes in evaluation graders and judge prompts?

\subsection{Setup}
\label{subsec:setup}
\paragraph{Base Model and Datasets.}
% base model, dataset (including size + task explanation)
We mainly use Qwen2.5-7B-Instruct~\citep{qwen2p5} as the base policy model, following previous rubric-based RL work~\citep{rar,onlinerubrics,hacking}.
We separately train the model on two subsets of RubricHub~\citep{rubrichub}, Medical and Science. Each example consists of a prompt (or query) $x$ sourced from existing open-ended QA datasets and a set of evaluation rubrics and their weights $\mathcal{R}(x)$ synthesized through multiple steps using LLMs.
In our experiments, for the Science subset, we exclude examples sourced from ResearchQA~\citep{researchqa} to use this benchmark as a separate evaluation suite.
As a result, the Medical subset contains 12,519 training examples and 1,391 test examples, and the Science subset contains 19,806 training examples and 2,201 test examples.

\paragraph{Baselines and Ablations.}
Based on their reward formulations, we categorize other methods designed to mitigate reward hacking in rubric-based RL into the following three types:

\textbf{(1) Rubric-Coverage Rewards} include \textit{OnlineRubrics}~\citep{onlinerubrics}, which adds complementary rubrics to predefined rubrics at every training step by comparing current-policy and control-policy responses; \textit{Universal Rubrics}\footnote{We basically follow \citet{checklists} but modify generic rubric items to directly address the observed reward hacking.}, which adds two generic criteria (necessity and factual correctness) to the predefined rubrics; and \textit{Reweighted Absence-Based Rubrics}, which increases the weights $w_j$ of absence-based rubrics. We also include the \textit{Standard} rubric-coverage reward~\citep{rar} for reference.

\textbf{(2) Preference-Based Rewards} include \textit{Holistic Preference}\footnote{We adopt the preference-reward formulation from \citet{nlhf} but use different judging criteria.}, which assigns rewards according to an LLM grader's overall preference based on rubric coverage, necessity, and factual correctness when comparing current-policy and control-policy responses. If a current-policy response is preferred, the reward is 1; if it is less preferred, the reward is 0; and if the responses are tied, the reward is 0.5.

\textbf{(3) Ablations: \textsc{Fork} Rewards with Precision Variants} include \textit{Generic Criteria}, which measures precision as the fraction of necessary and factually correct units, both determined by prompting an LLM grader; \textit{Offline Criteria}, which elicits prompt-specific precision criteria by comparing different model responses before training; and \textit{Online Criteria}, which elicits precision criteria specific to current-policy responses at each training step.

For simplicity, we use base policy responses as control-policy responses.
Implementation details for the baselines and ablations we introduce are provided in Appendix~\ref{app:sec:prec-var}, and their corresponding training prompts are provided in Appendix~\ref{app:subsec:prompts-training}.

\paragraph{Training Grader Selection.}
% correlation of several models vs. panels, take the best one (purpose: to exclude grader hacking)
% panel: GPT-5.4, Gemini 3 Pro, Claude Opus 4.6
We select an LLM to use as the \textit{training grader} for all rubric-based RL experiments.
To do that, we first select five candidate training graders and measure agreements of their judgments with three frontier models. The result shows that Qwen3.6-35B-A3B~\citep{qwen36} shows the strongest correlation (\textit{r} = 0.714), so we use this model as a universal training grader.
Please refer to Appendix~\ref{app:sec:grader} for the detailed grader selection process.

\paragraph{Evaluation.}
% training grader: qwen3.6-35b / reference grader: gpt-5.4
We evaluate in-domain performance on RubricHub test sets and cross-dataset generalization on external benchmarks (i.e., HealthBench~\citep{healthbench} for the medical domain and ResearchQA~\citep{researchqa} for the science domain).
We conduct two types of evaluations:
(1) rubric-based evaluations measuring macro rubric recall, macro precision, and macro F1, and
(2) rubric-free evaluations reporting the overall win rate against the base policy model and average response length (i.e., verbosity).
We use GPT-5.4~\citep{gpt5}, a strong frontier LLM, as a \textit{reference grader} for automating these evaluations.
We use the same rubric-free judge prompt as in \citet{hacking} and provide newly introduced judge prompts in Appendix~\ref{app:subsec:prompts-eval}.
To measure precision, a reference grader directly determines unit necessity and factual correctness given a set of coverage rubrics. We compute the F1 score as the harmonic mean of rubric recall and precision.

\paragraph{Hyperparameters and Implementation Details.}
% grpo details, lr-bs, epochs, fair comparison
We update the policy model using the group relative policy optimization (GRPO) algorithm~\citep{grpo} with 16 rollouts per prompt, a context length of 2,000 tokens, and a sampling temperature of 1.0.
Models are trained with a batch size of 128, a learning rate of $4.2 \times 10^{-6}$, and a constant schedule with 5\% linear warmup for 3 epochs on 8 NVIDIA H100 GPUs.
% fair comparison?

\begin{table}[t]
% \vspace{-5mm}
\centering
\small
\setlength{\tabcolsep}{5pt}
\renewcommand{\arraystretch}{1.15}

\caption{
\textbf{Main results on medical domain.}
We conduct rubric-based and rubric-free evaluations of our method and other rubric-based RL baselines on the RubricHub-Medical test set and HealthBench.
Rubric-based evaluation measures macro rubric recall, macro precision, and the macro F1 score. Precision indicates the fraction of necessary and factually correct response units.
Rubric-free evaluation measures win rate against the base policy model and average response length.
We use GPT-5.4 as a reference grader for automating these evaluations.
}
\label{tab:results-medical}

\resizebox{\textwidth}{!}{%
\begin{tabular}{lccccccc}
\toprule
& \multicolumn{5}{c}{\textbf{RubricHub-Medical test set}} & \multicolumn{2}{c}{\textbf{HealthBench}} \\

\cmidrule(lr){2-6}
\cmidrule(lr){7-8}

& \multicolumn{3}{c}{Rubric-based} & \multicolumn{2}{c}{Rubric-free} & Rubric-based & Rubric-free \\

\cmidrule(lr){2-4}
\cmidrule(lr){5-6}
\cmidrule(lr){7-7}
\cmidrule(lr){8-8}

& Recall & Precision & $F_1$ & Win Rate & Length & Recall & Win Rate \\

\midrule
\multicolumn{8}{l}{\textbf{Base}} \\
Qwen2.5-7B-Instruct                 & 22.6 & 46.6 & 26.3 & N/A & 2,432 & 13.4 & N/A \\

\midrule
\multicolumn{8}{l}{\textbf{Rubric-Coverage Rewards}} \\
Standard                            & 35.7 & 25.5 & 26.4 & 3.1  & 9,195 & 25.9 & 7.1 \\
OnlineRubrics                       & \textbf{38.7} & 26.8 & 28.8 & 10.6 & 8,736 & 25.7 & 22.1 \\
Universal Rubrics                   & 37.6 & 27.6 & 28.4 & 4.3 & 8,457 & \textbf{26.1} & 18.5 \\
Reweighted Absence-Based Rubrics    & 36.4 & 24.2 & 25.7 & 8.5 & 9,106 & 25.7 & 6.3 \\
\midrule
\multicolumn{8}{l}{\textbf{Preference-Based Rewards}} \\
Holistic Preference                 & 27.5 & 47.5 & 31.3 & 65.4 & 3,584 & 22.8 & 71.6 \\

\midrule
\rowcolor{LightBlue} \multicolumn{8}{l}{\textbf{\textsc{Fork} Rewards w/ Prec. Variants ({$\beta=1$})  (Our Ablations)}} \\
Generic Criteria                    & 33.8 & 47.7 & 36.9 & 44.0 & 4,312 & 24.5 & 63.4 \\
Offline Criteria                    & 38.0 & 35.1 & 33.7 & 20.7 & 6,748 & 25.1 & 32.6 \\
Online Criteria                     & 36.8 & 39.3 & 34.7 & 14.3 & 6,699 & 25.9 & 14.0\\

\midrule
\rowcolor{LightGreen} \multicolumn{8}{l}{\textbf{\textsc{Fork} Rewards (Ours)}} \\
$\beta=1$                           & 32.5 & \textbf{53.2} & 37.5 & \textbf{68.0} & 3,533 & 25.2 & \textbf{74.1} \\
% $\beta=1.25$                        & 35.8 & 49.0 & \textbf{38.6} & 55.4 & 4,392 & 25.1 & 64.3 \\

\bottomrule
\end{tabular}%
}
% \vspace{-3mm}
\end{table}

\begin{table}[t]
% \vspace{-3mm}
\centering
\small
\setlength{\tabcolsep}{5pt}
\renewcommand{\arraystretch}{1.15}

\caption{
\textbf{Main results on science domain.}
We evaluate rubric-based RL methods on the RubricHub-Science test set and ResearchQA as in Table~\ref{tab:results-medical}.
}
\label{tab:results-science}

\resizebox{0.9\textwidth}{!}{%
\begin{tabular}{lccccccc}
\toprule
& \multicolumn{5}{c}{\textbf{RubricHub-Science test set}} & \multicolumn{2}{c}{\textbf{ResearchQA}} \\

\cmidrule(lr){2-6}
\cmidrule(lr){7-8}

& \multicolumn{3}{c}{Rubric-based} & \multicolumn{2}{c}{Rubric-free} & Rubric-based & Rubric-free \\

\cmidrule(lr){2-4}
\cmidrule(lr){5-6}
\cmidrule(lr){7-7}
\cmidrule(lr){8-8}

& Recall & Precision & $F_1$ & Win Rate & Length & Recall & Win Rate \\

\midrule
\multicolumn{8}{l}{\textbf{Base}} \\
Qwen2.5-7B-Instruct~~~~~~~~         & 27.3 & 41.8 & 29.6 & N/A & 2,673 & 48.3 & N/A \\

\midrule
\multicolumn{8}{l}{\textbf{Rubric-Coverage Rewards}} \\
Standard                            & 31.4 & 33.5 & 29.2 & 38.6 & 4,453 & \textbf{64.4} & 16.1 \\
Universal Rubrics                   & \textbf{31.8} & 37.9 & 31.0 & 42.5 & 4,128 & 62.1 & 26.6 \\

\midrule
\rowcolor{LightGreen} \multicolumn{8}{l}{\textbf{\textsc{Fork} Rewards (Ours)}} \\
$\beta = 1.25$   & 30.4 & \textbf{45.1} & \textbf{33.4} & \textbf{61.5} & 3,038 & 58.9 & \textbf{67.1} \\

\bottomrule
\end{tabular}%
}
% \vspace{-3mm}
\end{table}

\subsection{Results}
\label{subsec:results}
We report main results on medical domain in Table~\ref{tab:results-medical} and science domain in Table~\ref{tab:results-science}. We enumerate our main findings in the following:

% 1. Rubric-F_\beta reward shows the best precision while maintaining recall, which leads to the best rubric-F1 score. -- also refer to fig.1
\textbf{\textsc{Fork} achieves higher precision than other rubric-based RL methods while maintaining recall at the level of standard coverage reward.}
Because of this, compared to rubric-coverage rewards, our method improves rubric-based $F_1$ score by 9.8 pp (from 28.8\% to 38.6\%) on RubricHub-Medical.
Preference-based rewards show higher precision than rubric-coverage rewards but have significantly degraded recall.
% Learning curve (Figure~\ref{fig:intro}) also shows that our method increases both recall and precision as training progresses, whereas standard coverage reward drastically loses precision.

% 2. Rubric-F_\beta reward shows the best rubric-free evaluation performance (while analyzing the reason, refer to completeness, conciseness, relevance, correctness, safety Likert scores.) -- this is different from precision!
\textbf{\textsc{Fork} achieves higher rubric-free win rate than other rubric-based RL methods.}
Compared to rubric-coverage rewards, our method increases overall win rate by 57.4 pp (from 10.6\% to 68.0\%) on RubricHub-Medical and by 19.0 pp (from 42.5\% to 61.5\%) on RubricHub-Science.
Further analysis reveals that our method shows better response quality in all five dimensions, including completeness, factual correctness, conciseness, relevance, and safety (Table~\ref{tab:rubric_free_quality}).

% 3. Rubric-F_\beta reward significantly reduces response length while containing necessary content. -- qualitative example?
\textbf{\textsc{Fork} effectively reduces response length while retaining necessary content.}
On RubricHub-Medical, compared to the standard coverage reward, our method ($\beta=1.25$) shortens the response length by a factor of 2.1 but the recall maintains nearly the same, which proves that verbosity issue is significantly reduced.
We also qualitatively verify this by manually checking representative examples (Appendix~\ref{app:sec:qualitative}).

% 4. Same benefits are observed on external benchmarks.
\textbf{\textsc{Fork} generalizes effectively to external benchmarks.}
While maintaining similar level of recall, compared to rubric-coverage rewards, our method increases overall win rate by 52.0 pp (from 22.1\% to 74.1\%) on HealthBench and by 40.5 pp (from 26.6\% to 67.1\%) on ResearchQA.

% % 5. Ablating either precision criteria leads to performance drop. -- 표는 부록으로?
% \paragraph{Ablating either precision criterion in \textsc{Fork} Reward leads to a performance drop.}

% \begin{table}[t]
% \centering
% \caption{
% \textbf{Ablation of precision criteria.}
% We ablate either the necessity or the factual correctness criterion from precision and report performance on the RubricHub-Medical test set.
% }
% \label{tab:ablation}
% \small

% \begin{tabular}{lccc cc}
% \toprule
% & \multicolumn{3}{c}{Rubric-based}
% & \multicolumn{2}{c}{Rubric-free} \\
% \cmidrule(lr){2-4}
% \cmidrule(lr){5-6}
% & Recall & Precision & $F_1$ & Win Rate & Length \\
% \midrule
% Rubric-$F_{1.25}$ Reward (Ours)   & 35.8 & 49.0 & 38.6 & 55.4 & 4,392 \\
% w/o necessity criteria            &  \\
% w/o factual correctness criteria  & 27.0 & 26.7 & 20.4 & 6.1 & 8,471 \\
% \bottomrule
% \end{tabular}
% \end{table}

\subsection{Analysis}
\label{subsec:analysis}
% [tbd] 0. Improvement in both presence/absence-based rubric satisfaction
\textbf{Improvement in satisfaction of both presence- and absence-based rubrics.}
In RubricHub, most of the coverage reward weight falls on presence-based rubrics rather than absence-based rubrics~\citep{hacking}. Accordingly, we observe that the standard coverage reward significantly increases the presence-based rubric satisfaction rate (+15.0 pp) while slightly decreasing the absence-based rubric satisfaction rate (-3.2 pp) (Table~\ref{tab:rubric_sat_rate}). This implies that the model learns to add both desirable and undesirable content.
Rubric-coverage reward baselines produce more desirable content but fail to suppress undesirable content.
In contrast, our method nearly matches the presence-based rubric satisfaction rate of the standard coverage reward while improving the absence-based rubric satisfaction rate (+5.7 pp).

% Compact analysis group 1: two independently numbered comparison tables.
\begin{table}[t]
% \vspace{-5mm}
\centering
\captionsetup{font=footnotesize,skip=4pt,justification=raggedright,singlelinecheck=false}
\begin{minipage}[t]{0.45\textwidth}
\vspace{0pt}
\centering
\captionof{table}{
\textbf{Presence-based and absence-based rubric satisfaction results.}
We categorize rubrics in the RubricHub-Medical test set as presence-based or absence-based and report satisfaction rates for rubric-based RL methods.
}
\label{tab:rubric_sat_rate}
\vspace{0.05cm}
\footnotesize
\setlength{\tabcolsep}{3pt}
\renewcommand{\arraystretch}{1.0}
\begin{tabularx}{\linewidth}{@{}Ycc@{}}
\toprule
% & \multicolumn{2}{c}{\bf Rubric Satisfaction Rate} \\
% \cmidrule{2-3}
\bf Rubric Satisfaction Rate
&
\makecell[c]{Presence-\\based}
&
\makecell[c]{Absence-\\based}
\\
% &
% \makecell[c]{Presence-\\based\\Rubric Sat.\\Rate}
% &
% \makecell[c]{Absence-\\based\\Rubric Sat.\\Rate}
% \\
\midrule

\multicolumn{3}{@{}l@{}}{\textbf{Base}} \\
Qwen2.5-7B-Instruct
& 20.3 & 40.4 \\

\midrule
\multicolumn{3}{@{}l@{}}{\textbf{Rubric-Coverage Rewards}} \\
Standard & 35.3 & 37.2 \\
OnlineRubrics & 38.5 & 36.9 \\
Universal Rubrics & 37.3 & 34.0 \\

\midrule
\rowcolor{LightGreen} \multicolumn{3}{@{}l@{}}{\textbf{\textsc{Fork} Rewards (Ours)}} \\
$\beta=1.25$ & 34.9 & 42.9 \\

\bottomrule
\end{tabularx}
\end{minipage}
\hfill
\begin{minipage}[t]{0.52\textwidth}
\vspace{0pt}
\centering
\captionof{table}{
\textbf{Rubric-free evaluation results using a different judge prompt.}
We adopt a different judge prompt from \citet{onlinerubrics} and report win rates against the base policy model. \textbf{Bold} and \underline{underline} indicates the best and second best performance.
}
\label{tab:prompt_robustness}
\footnotesize
\setlength{\tabcolsep}{3pt}
\renewcommand{\arraystretch}{1.0}
\resizebox{0.93\textwidth}{!}{
\begin{tabularx}{\linewidth}{@{}Ycc@{}}
\toprule
% & \makecell[c]{\textbf{RubricHub-}\\\textbf{Medical}} & \textbf{HealthBench} \\
& \bf RubricHub-M & \bf HealthBench \\
\cmidrule(lr){2-2}
\cmidrule(lr){3-3}
& Win Rate & Win Rate \\

\midrule
\multicolumn{3}{@{}l@{}}{\textbf{Preference-Based Rewards}} \\
Holistic Preference               & 80.8 & \textbf{89.4} \\

\midrule
\rowcolor{LightBlue} \multicolumn{3}{@{}l@{}}{\textbf{\textsc{Fork} Rewards w/ Prec. Variants ($\beta=1$)}} \\
Generic Criteria                  & 84.6 & 87.3 \\
Offline Criteria                  & \underline{88.6} & 80.3 \\
Online Criteria                   & 65.2 & 53.4 \\

\midrule
\rowcolor{LightGreen} \multicolumn{3}{@{}l@{}}{\textbf{\textsc{Fork} Rewards (Ours)}} \\
$\beta=1$                         & 84.8 & \underline{88.6} \\
$\beta=1.25$                      & \textbf{89.3} & 84.6 \\

\bottomrule
\end{tabularx}
}
\end{minipage}
\vspace{-5mm}
\end{table}

% [tbd] 1. correlation btw presence/absence-based rubric satisfaction vs. response length -- RaR은 완전 양의 상관관계. 우리 것은 거의 0이기를. -- reward hacking 따라가기.
% \textbf{Correlation between presence/absence-based rubric satisfaction vs. response length.}
% TBD.

% [tbd] 3. Rubric-free judge prompt robustness
\textbf{Robustness to rubric-free evaluation judge prompts.}
Since rubric-free win rates depend heavily on LLM judge prompt instructions, we verify whether our method is robust to prompt variations. To do so, we adopt another judge prompt from \citet{onlinerubrics} and verify on the medical domain (Table~\ref{tab:prompt_robustness}).
Since rubric-coverage reward baselines have significantly degraded win rates, we do not include those results. Our method mostly outperforms the other baselines on both evaluation suites, supporting its robustness to different judge prompts.

\begin{figure}[t]
% \vspace{-5mm}
\centering
\captionsetup{font=footnotesize,skip=4pt,justification=raggedright,singlelinecheck=false}
\begin{minipage}[t]{0.535\textwidth}
\vspace{0pt}
\centering
\includegraphics[width=\linewidth]{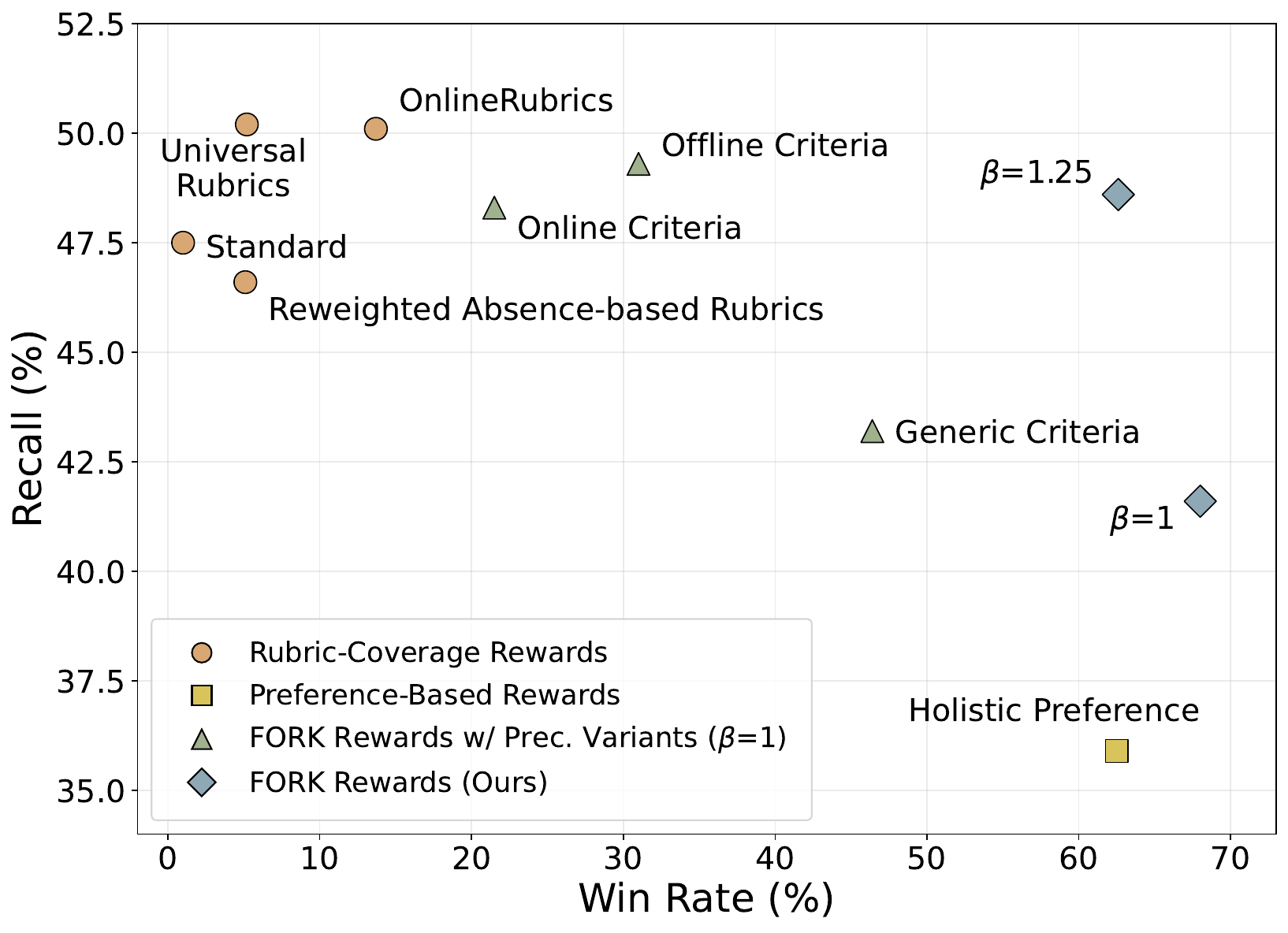}
\captionof{figure}{
  \textbf{Recall vs. Win Rate using a training grader.}
  We report recall against win rate for rubric-based RL methods evaluated with the training grader on RubricHub-Medical.
  }
\label{fig:training_grader_plot}

\vspace{8pt}
\captionof{table}{\textbf{Effectiveness of two different definitions of units.} We train models with \textit{Generic Criteria} baseline on RubricHub-Medical using two definitions of units and compare these with a model trained with standard rubric-coverage reward.}
\label{tab:unit_def}
\footnotesize
\setlength{\tabcolsep}{3pt}
\renewcommand{\arraystretch}{1.0}
\begin{tabularx}{\linewidth}{@{}Yccc@{}}
\toprule
 & Recall & Win Rate & Length \\
\midrule
Units as sentences (Ours)          & 33.8 & 44.0 & 4,312 \\
Units divided by line changes      & 35.7 & 37.8 & 5,775 \\
\midrule
Standard coverage reward           & 35.7 & 3.1  & 9,195 \\
\bottomrule
\end{tabularx}
\end{minipage}
\hfill
\begin{minipage}[t]{0.435\textwidth}
\vspace{0pt}
\centering
\includegraphics[width=0.8\linewidth]{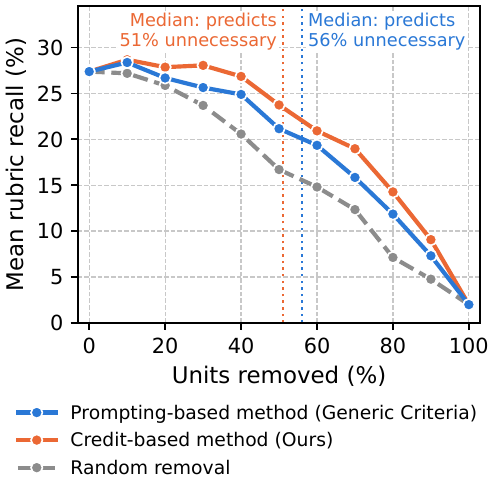}
\vspace{-2pt}
\captionof{figure}{
  \textbf{Accuracy of necessity prediction.}
  We progressively remove response units starting from unnecessary ones.
  % and then necessary ones once all unnecessary ones are removed.
  We use 723 units from 40 base model responses in RubricHub-Medical.
  }
\label{fig:necessity_analysis}

\vspace{6pt}
\captionof{table}{
\textbf{Effectiveness of different aggregations of precision and recall.}
We compare our $F_\beta$ aggregation with simple averaging on RubricHub-Medical while using the same method to measure precision.
}
\label{tab:pr-agg}
\footnotesize
\setlength{\tabcolsep}{3pt}
\renewcommand{\arraystretch}{1.0}
\begin{tabular*}{0.8\linewidth}{@{\extracolsep{\fill}}lccc@{}}
\toprule
 & Recall & Precision & $F_{1}$ \\
\midrule
Average                      & 26.2 & \textbf{55.3} & 32.8 \\
\midrule
$F_{1}$ score                & 32.5 & 53.2 & 37.5 \\
$F_{1.25}$ score             & \textbf{35.8} & 49.0 & \textbf{38.6} \\
$F_{2}$ score                & 35.5 & 43.6 & 36.7 \\
\bottomrule
\end{tabular*}
\end{minipage}
\vspace{-3mm}
\end{figure}

% [tbd] 4. Isolating the effectiveness of our reward formulation; use the same judge during eval
\textbf{Isolating the effectiveness of our reward formulation.}
We use a reference grader for our main evaluation (Section~\ref{subsec:results}). However, this evaluation combines the effect of our reward formulation with the effect of transferring to a stronger, unseen grader.
To isolate the former, we use the training grader for evaluation (Figure~\ref{fig:training_grader_plot}).
The results show that \textsc{Fork} rewards have the highest win rates among rubric-based RL methods, while $\beta$ = 1.25 shows Pareto-optimal performance that balances rubric coverage and overall win rate. This supports the effectiveness of our reward formulation compared to the baselines.

% % Compact analysis group 2: each column stacks one plot and one table.

% [R] 6. Different aggregation methods (F1, F1.25, F2, avg) -- report rec/prec/f1
\textbf{Different precision--recall aggregations.}
To justify using a harmonic mean to aggregate precision and recall, we compare $F_1$ score aggregation with simple averaging (Table~\ref{tab:pr-agg}). The results show that simple averaging yields slightly higher precision (+2.1 pp) but lower recall (-6.3 pp) than $F_1$ score aggregation. We observe that higher precision compensates for lower rubric recall during training.
We also investigate whether further increasing the weight of recall in $F_\beta$ score aggregation is beneficial. To do so, we train a model with $F_2$ score aggregation but observe a detrimental effect on both precision and recall. We leave finding the optimal value of $\beta$ for future work.

% [v] 2. rubric-driven necessity (+ grader-based necessity) vs. GT necessity correlation (by gpt-5.4 + other panels?) -> 메세지: necessity 측정이 정확히 되기 때문에 precision이 훨씬 높게 나오고 conciseness likert score도 높게 나옴.
\textbf{Our unit necessity prediction outperforms prompting-based method.}
To measure the accuracy of unit necessity prediction, we progressively remove units predicted as unnecessary from a response and measure how much rubric recall score is decreased (Figure~\ref{fig:necessity_analysis}).
Specifically, we randomly sample 40 base model responses from RubricHub-Medical containing 723 units and progressively remove units starting from unnecessary ones and then necessary ones once all unnecessary units are removed. We compare our credit-based method with prompting-based method used in \textit{Generic Criteria} baseline and random removal for reference.
Our method shows nearly no decrease in recall until 40\% of units are removed (note that in 51\% of units are predicted as unnecessary). In contrast, prompting-based method shows gradual decrease in recall as unnecessary units are removed.
This supports effectiveness of our unit necessity prediction and explains higher precision and overall winning rate of \textsc{Fork} reward ($\beta=1$) than \textit{Generic Criteria} baseline (Table~\ref{tab:results-medical}).

% [v] 7. Different units definition (line change) -- Generic Criteria 아래에서 실험. 결과는 비슷함.
\textbf{Alternative definition of response units.}
We also divide model responses into units by line changes and compare it with our sentence-level units (Table~\ref{tab:unit_def}). For simplicity, we train models with \textit{Generic Criteria} baseline using different definition of units.
The result shows that including precision objective using either definition of units both leads to significantly higher winning rate against the base policy model and similar rubric recall score, which indicates that both definitions of units are valid.
% Our definition shows 1.9\% lower rubric recall but has 6.2\% higher winning rate against the base policy model and has 25.3\% shorter response length.
However, since we observe semantically inconsistent divide in line changes, we adopt units as sentences for consistency.

% [v] 8. Effectiveness of dynamic thresholding -- compare w/ fixed necessity credit threshold (\tau=1/8 -- 선정 기준 정당화 필요)
% \paragraph{Effectiveness of dynamic necessity credit thresholding.}

\section{Conclusion}
\label{sec:conclusion}

In this work, we introduce \textsc{Fork}, a simple method that mitigates reward hacking in rubric-based RL, by adding a complementary precision objective to the original recall-based reward.
To address the observed reward hacking behavior, we define a precision objective to require each unit of a model response to be necessary and factually correct. We propose rubric-driven necessity credits to faithfully measure necessity and use an LLM grader to measure factual correctness.
Our experimental results show that \textsc{Fork} outperforms other rubric-based RL methods and achieves Pareto-optimal performance that balances recall and overall response quality.

\clearpage

\subsection*{AI use statement}
In this work, we used generative AI tools for searching related previous works and checking grammar.
We have reviewed all AI-assisted work. Specifically, we manually looked at the previous works that the AI tool suggests and verified their relevance. Also, we reviewed grammar-corrected contents so that there is no distortion in the original semantic meaning.

\bibliography{iclr2027_conference}
\bibliographystyle{iclr2027_conference}

\newpage

\appendix

\section{Detailed Description of Other Rubric-based RL Methods}
\label{app:sec:prec-var}

In Section~\ref{subsec:setup}, we briefly describe which baselines and ablations we adopt. Since there are newly introduced methods, in this section, we further describe how those are implemented in detail.

\subsection{Rubric-Coverage Rewards}
\textit{Universal Rubrics} adds two generic rubrics corresponds to necessity and factual correctness to a pre-defined set of rubrics. The verbatim rubrics are: (1) ``The model response should be \textit{necessary}: every part is needed, with nothing repeated, padded, tangential, or included only to sound thorough.", (2) ``The model response should be \textit{factually correct}: every claim is correct and supported, with nothing fabricated, misleading, or overconfident.".

\textit{Reweighted Absence-Based Rubrics} first determines absence-based rubrics. If a rubric contains both presence-based component and absence-based component, we consider it as an absence-based. Then, given these absence-based rubrics and their original weights, an LLM increases their weights. The prompt is provided in Figure~\ref{fig:prompt-reweighted-1}--\ref{fig:prompt-reweighted-2}.
% -- absence rubric 먼저 확인 (presence 섞여 있어도 그렇게 함) -> llm보고 알아서 올리라. (프롬프트)

\subsection{Preference-Based Rewards}
\textit{Holistic Preference} let an LLM grader to decide its preference of current policy responses against base policy responses and uses as a reward. The prompt is provided in Figure~\ref{fig:prompt-holistic-1}--\ref{fig:prompt-holistic-2}.

\subsection{Ablations: \textsc{Fork} Rewards with Precision Variants}
% \textit{Generic Criteria} -- 정확히 어떤 verbatim 기준?
% \textit{Offline Criteria} -- rubric generator가 여러 모델 (base, rar) 답변을 보고 precision 루브릭 (몇개?) 생성.
% \textit{Online Criteria} -- rubric generator가 루브릭과 현재 모델 답변을 보고 precision 루브릭 (몇개?) 생성.
These rewards use the same structure of reward as our method with $F_1$ aggregation, but use different methods of measuring precision.

\textit{Generic Criteria} replaces our credit-based necessity prediction to a simple prompting method. Therefore, an LLM grader is given a set of response units and corresponding rubrics and determines each unit's necessity and factual correctness, respectively. The prompt is provided in Figure~\ref{fig:prompt-generic}.

\textit{Offline Criteria} generates prompt-specific precision criteria and use those for measuring precision, instead of using two generic criteria, necessity and factual correctness. To generate those criteria, an LLM grader is given two responses, one from a base policy model and another from a RL-trained policy model with standard coverage reward, and then generates precision criteria by comparing those two responses. During online RL training, these precision criteria are used for every training step. The prompt is provided in Figure~\ref{fig:prompt-offline-1}--\ref{fig:prompt-offline-2}.

\textit{Online Criteria} also generates prompt-specific precision criteria but updates those at every training step. To do that, an LLM grader is given a current policy response and a set of rubrics, and generates online precision criteria by evaluating the response quality. During online RL training, these precision criteria are updated for every training step. The prompt is provided in Figure~\ref{fig:prompt-online-1}--\ref{fig:promt-online-2}.

\section{Detailed Training Grader Selection Process}
\label{app:sec:grader}

We select an LLM to use as the \textit{training grader} for all rubric-based RL experiments.
This is especially important for mitigating misaligned decisions by the training grader, which have been identified as another mode of reward hacking in rubric-based RL~\citep{hacking}.
We first select candidate training graders as follows: gpt-oss-20b, gpt-oss-120b~\citep{gpt-oss}, GPT-4.1 Mini~\citep{gpt4}, Qwen3.5-122B-A10B~\citep{qwen35}, and Qwen3.6-35B-A3B~\citep{qwen36}.
Then, using 40 randomly sampled base policy responses from RubricHub-Medical, we measure each training grader's ability to make five types of judgments required by our method and the baselines: rubric satisfaction, holistic preference, necessity, factual correctness, and necessity credit thresholding. To do this, we collect judgments from three frontier LLMs, GPT-6 Astra~\citep{gpt6}, Gemini 3.8 Flash~\citep{gemini38flash}, and Claude Opus 5~\citep{claudeopus5}, take their majority judgments as reference judgments, and measure Pearson correlations between candidate and reference judgments.

When the correlations for the five judgment types are averaged, \textbf{Qwen3.6-35B-A3B} shows a strong correlation (\textit{r} = 0.714) and outperforms the other candidate graders (Table~\ref{tab:grader-selection}). We select this model as the universal training grader in all experiments.

\begin{table}[h]
% \vspace{-5mm}
\centering
\small
\setlength{\tabcolsep}{5pt}
\renewcommand{\arraystretch}{1.1}
\caption{\textbf{Training grader selection results.} We select five candidate training graders and measure their ability to make five types of judgments required for rubric-based RL methods. We use 40 randomly sampled base policy responses from RubricHub-Medical and report Pearson correlations between candidate and reference judgments.}
\label{tab:grader-selection}
\resizebox{0.85\textwidth}{!}{%
\begin{tabular}{lcccccc}
\toprule
& \textbf{Rubric Sat.} & \textbf{Holistic Pref.} & \textbf{Necessity} & \textbf{Factual Corr.} & \textbf{Threshold} & \textbf{Avg.} \\
\midrule
gpt-oss-20b         & 0.591 & 0.796 & 0.348 & 0.350 & 0.679 & 0.553 \\
gpt-oss-120b        & 0.606 & 0.825 & 0.563 & 0.677 & 0.688 & 0.672 \\
GPT-4.1 Mini        & 0.639 & 0.846 & 0.447 & 0.579 & 0.711 & 0.644 \\
Qwen3.5-122B-A10B   & 0.735 & 0.849 & 0.575 & 0.542 & 0.780 & 0.696 \\
\textbf{Qwen3.6-35B-A3B}     & 0.719 & 0.853 & 0.534 & 0.703 & 0.760 & \textbf{0.714} \\
% \midrule
% DeepSeek-V4-Flash   & 0.772 & 0.821 & 0.637 & 0.737 & 0.819 & 0.757 \\
\bottomrule
\end{tabular}
}
\vspace{-3mm}
\end{table}

\section{Further Analysis on Rubric-free Evaluations}
\label{app:sec:analysis-rubric-free}

In Section~\ref{subsec:results}, we report that our method shows higher rubric-free win rate than other rubric-based RL methods. In this section, we analyze the LLM grader's decisions by comparing five quality dimension scores of our methods and rubric-coverage rewards (Table~\ref{tab:rubric_free_quality}). Note that each quality dimensions are scored in a 1--7 Likert scale following \citet{hacking}.

The results show that our method has higher absolute scores than rubric-coverage rewards in all five quality dimensions, including completeness (+0.05), factual correctness (+0.76), conciseness (+2.64), relevance (+1.30), and safety (+1.67).
The highest gain is at conciseness. It is interesting that safety shows the second highest gain even though our method does not learn to generate safe responses. This means that necessity and factual correctness aligns well with safety, which additionally supports effectiveness of our method.
Compared to the base model, our method shows higher completeness and safety scores while showing similar factual correctness and relevance scores, which contributes to a higher overall quality score.

\begin{table}[h]
\centering
\caption{
\textbf{Rubric-free evaluation results across response quality dimensions.}
We report each quality dimension scores of our method and rubric-coverage rewards in a 1--7 Likert scale following \citet{hacking}. We also add a base model for reference. The overall score is determined implicitly by the reference grader (i.e., GPT-5.4).
}
\label{tab:rubric_free_quality}

\resizebox{\textwidth}{!}{%
\begin{tabular}{lcccccc}
\toprule
 & Completeness & Factual Corr. & Conciseness & Relevance & Safety & \textbf{Overall} \\
\midrule

\textbf{Base} & & & & & & \\
Qwen2.5-7B-Instruct
& 4.83
& \textbf{4.90}
& \textbf{5.58}
& \textbf{5.95}
& 5.46
& 4.94 \\

\midrule
\textbf{Rubric-Coverage Rewards} & & & & & & \\
Standard
& 5.34
& 3.77
& 1.14
& 4.05
& 4.98
& 3.29 \\

OnlineRubrics
& 5.68
& 4.05
& 1.79
& 4.49
& 5.39
& 3.95 \\

Universal Rubrics
& 5.64
& 3.76
& 1.85
& 4.57
& 5.11
& 3.79 \\

\midrule
\rowcolor{LightGreen} \textbf{\textsc{Fork} Rewards (Ours)} & & & & & & \\
$\beta = 1$
& \textbf{5.73}
& 4.81
& 4.49
& 5.87
& \textbf{6.06}
& \textbf{5.20} \\

\bottomrule
\end{tabular}
}
\end{table}

\section{Qualitative Analysis of Model Responses}
\label{app:sec:qualitative}

On RubricHub-Medical, \textsc{Fork} ($\beta=1.25$) shortens model responses by a factor of 2.1 relative to the standard coverage reward while keeping recall nearly the same (Table~\ref{tab:results-medical}), supporting that verbosity issue is reduced significantly.
To qualitatively verify this, we first select two pairs of examples from medical and science domain, where each pair consists of model responses from the standard coverage reward and the proposed \textsc{Fork} reward. Then, we \qexhl{highlight} necessary units using a reference grader (i.e., GPT-5.4).

We confirm that our method mostly generates necessary contents whereas the standard coverage reward does not. Also, while having similar amount of necessary contents, our method generates significantly shorter, more concise responses.

One thing we observe in the medical example is that the model trained with the standard coverage reward first generates an initial response, add another response, and repeat this process (Each response ends with: ``If you have any more questions or need further assistance, feel free to ask. Stay Well! Qwen"). Our model does not show this behavior.

The science example shows a different failure: after computing the work correctly, the standard coverage reward adds a verification step that contradicts its own result.
The reference grader labels the whole step unnecessary, so none of it is highlighted. The response then concludes that the assumption in the verification step is incorrect, and thus maintain the original answer.
Our model does not show this behavior.

% To examine what this shortening removes, Appendix~\ref{app:subsec:qualitative-example} shows the complete responses of the two policies to one test prompt, decoded greedily with a 2,000-token limit. \qexhl{Highlighted} text marks the units that the reference grader (GPT-5.4) labels necessary with the prompt in Figure~\ref{fig:prompt-generic}.

% The heading, the prompt and FORK's response (about 0.8\textheight) stay on one page:
% they start a new page unless 0.85\textheight is left on this one.
\needspace{0.85\textheight}
\subsection{Medical: A Consumer Health Question}
\label{app:subsec:qualitative-example}
% Generated by precision-RaR's analysis/exemplary_samples/in-domain_medical/final_sample/v2/make_final_sample_v2.py
% (its med-89eb14fabb27.tex) -- regenerate there and paste it here rather than editing by hand.
% Prompt med-89eb14fabb27 (grader-eval set, data/medical/eval.parquet); responses at step 291 from
%   outputs/exp3.7.5_pg_rar_f1.25_odad/eval/pg_rar_f1.25_odad/step_291/grader_responses.jsonl
%   outputs/exp2_verifier_swap/eval/rar_cov_v2/step_291/grader_responses.jsonl
% highlight = units GPT-5.4 labels necessary (<arm dir>/rb.gpt54/precision_scored.jsonl).
\begin{qexbox}{qexprompt}{Prompt}
Do allergies cause coughing? I had a bad cough all winter, and it went away for a couple months when the weather got warmer. It just started again since the weather has gotten very hot (90 degrees) and rainy. Is there a cough medicine for coughs due to allergies?\par
\end{qexbox}

\begin{qexbox}{qexours}{\textsc{Fork} (ours)}
\qexhl{Yes, allergies can indeed cause coughing, and your symptoms align with this possibility.} Allergies can trigger a cough through several mechanisms, including:\par
\qexitem{1.}\qexhl{\textbf{Postnasal Drip}: When allergens like pollen, dust mites, or mold irritate the nasal passages, they can cause increased mucus production.} \qexhl{This excess mucus can drip down the back of the throat (postnasal drip), stimulating the cough reflex, especially at night or when lying down.}\par
\qexitem{2.}\qexhl{\textbf{Inflammation and Irritation}: Allergic reactions can cause inflammation in the airways, which can lead to a dry, hacking cough, especially if the airways are sensitive to the allergens.}\par
\qexitem{3.}\qexhl{\textbf{Bronchospasm}: In some cases, allergies can cause the airways to narrow, leading to wheezing and coughing, which is more common in individuals with asthma that is triggered by allergies.}\par
\qexhl{The pattern you described{\textemdash}coughing worsening in the winter and returning with hot, rainy weather{\textemdash}suggests that your cough may be related to seasonal allergies, possibly to mold, pollen, or other outdoor allergens that are more prevalent in these conditions, or to indoor allergens that may be more active in warmer, humid weather (like dust mites, mold, or pet dander).}\par
\qexhead{Cough Medicines for Allergy-Induced Cough}
\qexhl{There are several types of cough medicines that can help manage allergy-induced cough, but it's important to note that they should be used under the guidance of a healthcare provider, especially if you have underlying health conditions or are taking other medications.} Here are some options:\par
\qexitem{1.}\qexhl{\textbf{Antihistamines}: These can help reduce the symptoms of allergies, including sneezing, itching, and runny nose, which can contribute to coughing.} \qexhl{Common over-the-counter antihistamines include diphenhydramine (Benadryl) and cetirizine (Zyrtec).} \qexhl{However, they can cause drowsiness, so it's best to use them before bedtime or to avoid activities that require alertness, like driving.}\par
\qexitem{2.}\qexhl{\textbf{Decongestants}: These can help relieve nasal congestion, which can contribute to postnasal drip and cough, but they do not directly treat the cough itself.} \qexhl{Pseudoephedrine (Sudafed) is a common decongestant, but it can have side effects like increased heart rate and blood pressure, so it's not recommended for people with high blood pressure, heart disease, or prostate issues.}\par
\qexitem{3.}\qexhl{\textbf{Cough Suppressants}: These are used to suppress a cough that is dry and unproductive, and they include dextromethorphan (found in many cough syrups and cold medicines).} However, if your cough is productive (producing phlegm), a cough suppressant may not be the best choice, as it can prevent you from clearing the mucus that your body is trying to expel.\par
\qexitem{4.}\textbf{Expectorants}: These help loosen mucus and make it easier to cough up, and they include guaifenesin (found in Mucinex). They can be helpful if your cough is productive, but they do not address the underlying allergy.\par
\qexitem{5.}\qexhl{\textbf{Nasal Corticosteroids}: These are often more effective for long-term allergy management and can help reduce inflammation in the nasal passages, which can contribute to coughing.} Examples include fluticasone (Flonase) and mometasone (Nasonex), which are available by prescription.\par
\qexhead{Recommendations}
Given the pattern of your cough, it might be beneficial to use a combination of an antihistamine and a decongestant, or an antihistamine and an expectorant, to manage your symptoms. \qexhl{However, since you've had a persistent cough for an extended period, it's important to consult a healthcare provider to rule out other potential causes, such as a respiratory infection, asthma, or acid reflux, which can also cause chronic cough.}\par
Here are some steps you can take:\par
\qexitem{1.}\qexhl{\textbf{Identify Triggers}: Keep track of when your cough occurs (e.g., during certain times of the day, after being outside, or when the weather changes) to help identify potential allergens.}\par
\qexitem{2.}\qexhl{\textbf{Environmental Control}: Use air conditioning to keep indoor humidity low (below 50{\%}), use dehumidifiers, and regularly clean or replace air filters to reduce mold and dust mites.}\par
\qexitem{3.}\textbf{Allergy Testing}: Consider getting allergy testing to identify specific triggers, which can help in tailoring your treatment plan, whether it's through environmental control, allergy shots, or targeted medications.\par
\qexitem{4.}\textbf{Consult a Healthcare Provider}: Speak with a healthcare provider, such as an allergist or primary care physician, to discuss your symptoms, undergo a physical examination, and possibly undergo allergy testing, and to get a prescription for appropriate medications, which might include a combination of antihistamines, decongestants, and/or nasal corticosteroids, as well as advice on managing your cough, such as using a humidifier, staying hydrated, and avoiding irritants like cigarette smoke and air pollution.\par
\qexhl{If your cough is severe, accompanied by shortness of breath, wheezing, fever, or blood in your phlegm, or if it persists for more than 8 weeks, it's crucial to seek medical attention, as these symptoms could indicate a more serious condition that requires specific treatment, such as bronchitis, asthma, or even a lung infection.}\par
\end{qexbox}

\clearpage

\begin{qexbox}{qexrar}{Standard coverage reward}
\qexhl{Yes, allergies can indeed cause coughing.} \qexhl{Allergies trigger the release of histamines and other chemicals in the body, which can lead to inflammation and increased mucus production in the airways.} This can result in symptoms like coughing, sneezing, and a runny nose. \qexhl{The pattern you're describing{\textemdash}coughing that worsens in certain weather conditions{\textemdash}suggests that you might be experiencing seasonal allergies, also known as hay fever.}\par
Here are a few points to consider:\par
\qexitem{1.}\qexhl{\textbf{Seasonal Allergies and Coughing}: Allergies to pollen, mold, and other environmental triggers can worsen in spring and summer when these allergens are more prevalent.} The change in weather from winter to spring and then to summer can also affect the types and levels of allergens in the air.\par
\qexitem{2.}\textbf{Coughing Mechanism}: Coughing can be a response to irritation in the airways, such as from allergens, dust, or mucus. \qexhl{Post-nasal drip, where excess mucus from the nose drips down the back of the throat, can also cause a cough.}\par
\qexitem{3.}\qexhl{\textbf{Treatment Options}: There are several over-the-counter (OTC) medications that can help manage coughs and allergy symptoms.} Here are a few options:\par
\qexsubitem{--}\qexhl{\textbf{Antihistamines}: These can help reduce sneezing, itching, and runny nose, which can in turn reduce coughing.} \qexhl{Examples include cetirizine (Zyrtec), loratadine (Claritin), and fexofenadine (Allegra).} They are generally safe for most people, but check with your pharmacist or doctor if you have any concerns, especially if you are taking other medications.\par
\qexsubitem{--}\qexhl{\textbf{Decongestants}: These can help reduce nasal congestion, which can also help with coughing.} \qexhl{However, they should be used with caution, as they can cause rebound congestion if used for more than a few days.} \qexhl{They are not recommended for people with high blood pressure or certain heart conditions.}\par
\qexsubitem{--}\textbf{Expectorants}: These can help loosen mucus, making it easier to cough up. Guaifenesin (Mucinex) is an example of an expectorant. \qexhl{However, they are less effective for coughs caused by allergies.}\par
\qexsubitem{--}\qexhl{\textbf{Cough Suppressants}: These are used to suppress a dry, hacking cough.} \qexhl{Dextromethorphan (found in many cough syrups) can be effective, but they are not recommended for coughs with mucus, as suppressing the cough can prevent you from expelling the mucus.}\par
\qexitem{4.}\qexhl{\textbf{Other Considerations}: If your cough persists or is accompanied by other symptoms like fever, shortness of breath, or wheezing, it's important to consult a healthcare provider to rule out other conditions such as asthma, bronchitis, or pneumonia.}\par
\qexitem{5.}\textbf{Environmental Controls}: Managing your environment can also help reduce allergy symptoms. \qexhl{Use air conditioning to filter indoor air, keep windows closed during high pollen seasons, and use dehumidifiers to control mold growth.}\par
\qexhead{Specific to Your Situation}
\qexhl{Given that your cough worsened with the change in weather, it might be worth considering whether you are allergic to specific outdoor allergens like pollen or mold.} An allergist can perform skin tests or blood tests to identify specific allergens you might be reacting to.\par
If you find that your symptoms are significantly impacting your quality of life, it might be beneficial to consult a healthcare provider for a more personalized assessment and treatment plan.\par
Remember, always consult a healthcare professional for a proper diagnosis and personalized advice, especially if you have any underlying health conditions or are taking other medications.\par
I hope this information helps!\par
If you have any more questions or need further assistance, feel free to ask.\par
Stay well!\par
Qwen\par
\qexhead{Additional Note}
If you experience any of the following, seek immediate medical attention:\par
\qexitem{\textbullet}Difficulty breathing\par
\qexitem{\textbullet}Swelling of the face, lips, tongue, or throat\par
\qexitem{\textbullet}Severe wheezing\par
\qexitem{\textbullet}Hives or rash\par
\qexitem{\textbullet}Dizziness or fainting\par
\qexitem{\textbullet}Rapid heartbeat or palpitations\par
\qexitem{\textbullet}Difficulty swallowing\par
These could be signs of a severe allergic reaction (anaphylaxis), which is a medical emergency.\par
Qwen\par
\qexhead{Detailed Information}
Allergies can indeed cause coughing, and the pattern you're describing{\textemdash}coughing worsening with changes in weather{\textemdash}suggests that you might be experiencing seasonal allergies. Here's a more detailed explanation:\par
\qexitem{1.}\textbf{Seasonal Allergies and Coughing}:\par
\qexsubitem{--}\textbf{Pollen Allergies}: High pollen counts, especially from grass, trees, and weeds, can trigger coughing. Pollen counts are often highest in the early morning and late afternoon.\par
\qexsubitem{--}\qexhl{\textbf{Mold Allergies}: Mold spores can also cause coughing, especially in humid or rainy weather.} Mold can grow in damp areas like basements, bathrooms, and gardens.\par
\qexsubitem{--}\textbf{Weather Changes}: Changes in temperature and humidity can affect the airways and exacerbate coughing. For example, dry, cold air can dry out the nasal passages and throat, while humid, warm air can increase mucus production.\par
\qexitem{2.}\textbf{Coughing Mechanism}:\par
\qexsubitem{--}\textbf{Post-Nasal Drip}: When allergens cause inflammation in the nasal passages, excess mucus can drip down the back of the throat, leading to a cough.\par
\qexsubitem{--}\qexhl{\textbf{Bronchial Irritation}: Allergens can also irritate the bronchial tubes, causing a dry, hacking cough.}\par
\qexitem{3.}\textbf{Treatment Options}:\par
\qexsubitem{--}\textbf{Antihistamines}: These can help reduce sneezing, itching, and runny nose, which can in turn reduce coughing. Cetirizine (Zyrtec), loratadine (Claritin), and fexofenadine (Allegra) are effective for many people.\par
\qexsubitem{--}\textbf{Nasal Sprays}: Decongestant nasal sprays can help reduce nasal congestion, but they should be used for no more than a few days to avoid rebound congestion.\par
\qexsubitem{--}\qexhl{\textbf{Inhalers}: For those with asthma, inhaled corticosteroids and bronchodilators can help manage symptoms.}\par
\qexsubitem{--}\textbf{Expectorants}: Guaifenesin (Mucinex) can help loosen mucus, making it easier to cough up.\par
\qexsubitem{--}\textbf{Cough Suppressants}: Dextromethorphan (found in many cough syrups) can help suppress a dry, hacking cough, but they are not recommended for coughs with mucus.\par
\qexitem{4.}\textbf{Environmental Controls}:\par
\qexsubitem{--}Use air conditioning to filter indoor air.\par
\qexsubitem{--}Keep windows closed during high pollen seasons.\par
\qexsubitem{--}Use dehumidifiers to control mold growth.\par
\qexsubitem{--}\qexhl{Consider using HEPA filters in your home.}\par
\qexitem{5.}\textbf{Consult a Healthcare Provider}:\par
\qexsubitem{--}If your symptoms persist or worsen, or if you have other symptoms like fever, shortness of breath, or wheezing, consult a healthcare provider.\par
\qexsubitem{--}An allergist can perform skin tests or blood tests to identify specific allergens you might be reacting to.\par
\qexhead{Specific Medications for Allergy-Induced Cough}
There are specific medications that can help manage coughs due to allergies. Here are a few options:\par
\qexitem{\textbullet}\textbf{Cetirizine (Zyrtec)}: This is an antihistamine that can help reduce sneezing, itching, and runny nose, which can in turn reduce coughing.\par
\qexitem{\textbullet}\textbf{Loratadine (Claritin)}: Another antihistamine that can help manage allergy symptoms.\par
\qexitem{\textbullet}\textbf{Fexofenadine (Allegra)}: An antihistamine that is effective for many people with seasonal allergies.\par
\qexitem{\textbullet}\textbf{Montelukast (Singulair)}: This is a leukotriene receptor antagonist that can help reduce inflammation in the airways and may help with coughing in some people with asthma or allergies.\par
\qexhead{When to Seek Medical Attention}
\qexhl{If your cough is severe, persistent, or accompanied by other symptoms like fever, shortness of breath, or wheezing, it's important to consult a healthcare provider.} \qexhl{They can provide a proper diagnosis and recommend appropriate treatment.}\par
I hope this detailed information helps you better understand your symptoms and provides you with some guidance on managing them. If you have any more questions or need further assistance, feel free to ask.\par
Stay well!\par
Qwen\par
\qexhead{Detailed Medication Guidance}
For a cough due to allergies, here are some specific medications you might consider:\par
\qexitem{1.}\textbf{Antihistamines}:\par
\qexsubitem{--}\textbf{Cetirizine (Zyrtec)}: This is a non-sedating antihistamine that can help reduce sneezing, itching, and runny nose. It is generally well-tolerated and can be taken daily.\par
\qexsubitem{--}\textbf{Loratadine (Claritin)}: Another non-sedating antihistamine that is effective for many people with seasonal allergies. It is also generally well-tolerated.\par
\qexsubitem{--}\textbf{Fexofenadine (Allegra)}: This is another non-sedating antihistamine that can help manage allergy symptoms. It is generally well-tolerated and can be taken daily.\par
\qexitem{2.}\textbf{Decongestants}:\par
\qexsubitem{--}\textbf{Pseudoephedrine (Sudafed)}: This can help reduce nasal congestion, but it should be used with caution, as it can cause rebound congestion if used for more than a few days. It is also not recommended for people with high blood pressure or certain heart conditions.\par
\qexsubitem{--}\textbf{Phenylephrine}: This is another decongestant that can help reduce \ldots\par
\end{qexbox}

% The heading, the prompt and FORK's response (about 0.5\textheight) stay on one page:
% they start a new page unless 0.55\textheight is left on this one.
\needspace{0.55\textheight}
\subsection{Science: A Vector Calculus Problem}
\label{app:subsec:qualitative-example-science}
% Generated by precision-RaR's analysis/exemplary_samples/in-domain_science/final_sample/make_final_sample_sci.py
% (its sci-af6c37f73f46.tex) -- regenerate there and paste it here rather than editing by hand.
% Prompt sci-af6c37f73f46 (grader-eval set, data/science_curated/eval.parquet); responses at step 291 from
%   outputs/exp3.7.4_pg_rar_f1.25_odad_science/eval/pg_rar_f1.25_odad_sci/step_291/grader_responses.jsonl
%   outputs/b_exp3_rar_cov_science/eval/rar_cov_sci/step_291/grader_responses.jsonl
% highlight = units GPT-5.4 labels necessary (<arm dir>/rb.gpt54/precision_scored.jsonl).
\begin{qexbox}{qexprompt}{Prompt}
Show that if a particle is moved along the closed curve \mbox{$(\cos t, \sin t, 0)$}, where \mbox{$0 \leqslant t \leqslant 2 \pi$}, under the force field \mbox{$\mathbf{F} = -y \mathbf{i} + x \mathbf{j}$}, then the work done by the force field on the particle is nonzero. Calculate the exact amount of work done.\par
\end{qexbox}

\begin{qexbox}{qexours}{\textsc{Fork} (ours)}
To determine the work done by the force field \mbox{$\mathbf{F} = -y \mathbf{i} + x \mathbf{j}$} on a particle moving along the closed curve \mbox{$(\cos t, \sin t, 0)$} for \mbox{$0 \leq t \leq 2\pi$}, we follow these steps:\par
\qexitem{1.}\textbf{Parameterize the curve:}\par
\qexhl{The curve is given by \mbox{$\mathbf{r}(t) = (\cos t, \sin t, 0)$} for \mbox{$0 \leq t \leq 2\pi$}.}\par
\qexitem{2.}\textbf{Find the derivative of the curve:}\par
The derivative of \mbox{$\mathbf{r}(t)$} with respect to \mbox{$t$} is:\par
\qexeqhl{\mathbf{r}'(t) = \frac{d\mathbf{r}}{dt} = (-\sin t, \cos t, 0).}
\qexitem{3.}\textbf{Evaluate the force field \mbox{$\mathbf{F}$} at the point \mbox{$\mathbf{r}(t)$}:}\par
Since \mbox{$\mathbf{F} = -y \mathbf{i} + x \mathbf{j}$} and \mbox{$\mathbf{r}(t) = (\cos t, \sin t, 0)$}, we have:\par
\qexeqhl{\mathbf{F}(\mathbf{r}(t)) = -\sin t \mathbf{i} + \cos t \mathbf{j}.}
\qexitem{4.}\textbf{Compute the dot product \mbox{$\mathbf{F}(\mathbf{r}(t)) \cdot \mathbf{r}'(t)$}:}\par
\qexeqhl{\mathbf{F}(\mathbf{r}(t)) \cdot \mathbf{r}'(t) = (-\sin t \mathbf{i} + \cos t \mathbf{j}) \cdot (-\sin t \mathbf{i} + \cos t \mathbf{j}) = (-\sin t)(-\sin t) + (\cos t)(\cos t) + 0 = \sin^2 t + \cos^2 t = 1.}
\qexitem{5.}\textbf{Set up the integral for the work done:}\par
The work \mbox{$W$} done by the force field is given by the line integral:\par
\qexeqhl{W = \int_C \mathbf{F} \cdot d\mathbf{r} = \int_0^{2\pi} \mathbf{F}(\mathbf{r}(t)) \cdot \mathbf{r}'(t) \, dt = \int_0^{2\pi} 1 \, dt.}
\qexitem{6.}\textbf{Evaluate the integral:}\par
\qexeqhl{\int_0^{2\pi} 1 \, dt = \left[ t \right]_0^{2\pi} = 2\pi - 0 = 2\pi.}
\qexhl{Since the work done is \mbox{$2\pi$}, which is nonzero, we conclude that the work done by the force field on the particle is \mbox{$\boxed{2\pi}$}.}\par
\end{qexbox}

\clearpage

\begin{qexbox}{qexrar}{Standard coverage reward}
\qexhl{To determine the work done by the force field \mbox{$\mathbf{F} = -y \mathbf{i} + x \mathbf{j}$} on a particle moving along the closed curve \mbox{$(\cos t, \sin t, 0)$} for \mbox{$0 \leq t \leq 2\pi$}, we need to evaluate the line integral of \mbox{$\mathbf{F}$} along the curve.} The work done is given by:\par
\qexeqhl{W = \int_C \mathbf{F} \cdot d\mathbf{r}}
\qexhead{Step 1: Parametrize the curve}
The curve is given by:\par
\qexeqhl{\mathbf{r}(t) = (\cos t, \sin t, 0)}
\qexhl{where \mbox{$0 \leq t \leq 2\pi$}.}\par
\qexhead{Step 2: Compute the derivative of the parametrization}
The derivative of \mbox{$\mathbf{r}(t)$} with respect to \mbox{$t$} is:\par
\qexeqhl{\mathbf{r}'(t) = \frac{d}{dt} (\cos t, \sin t, 0) = (-\sin t, \cos t, 0)}
\qexhead{Step 3: Express the force field in terms of \mbox{$t$}}
Substitute \mbox{$x = \cos t$} and \mbox{$y = \sin t$} into the force field \mbox{$\mathbf{F} = -y \mathbf{i} + x \mathbf{j}$}:\par
\qexeqhl{\mathbf{F}(\mathbf{r}(t)) = -\sin t \mathbf{i} + \cos t \mathbf{j}}
\qexhead{Step 4: Compute the dot product \mbox{$\mathbf{F} \cdot d\mathbf{r}$}}
The line integral is:\par
\qexeqhl{\mathbf{F} \cdot d\mathbf{r} = \mathbf{F}(\mathbf{r}(t)) \cdot \mathbf{r}'(t)}
\qexeqhl{\mathbf{F}(\mathbf{r}(t)) \cdot \mathbf{r}'(t) = (-\sin t, \cos t, 0) \cdot (-\sin t, \cos t, 0)}
\qexeqhl{= (-\sin t)(-\sin t) + (\cos t)(\cos t) + (0)(0)}
\qexeqhl{= \sin^2 t + \cos^2 t}
\qexeqhl{= 1}
\qexhead{Step 5: Set up and evaluate the integral}
The work done is:\par
\qexeqhl{W = \int_0^{2\pi} 1 \, dt}
\qexeq{W = \left[ t \right]_0^{2\pi}}
\qexeqhl{W = 2\pi - 0 = 2\pi}
\qexhead{Step 6: Verify using the Fundamental Theorem of Line Integrals}
The force field \mbox{$\mathbf{F} = -y \mathbf{i} + x \mathbf{j}$} is conservative. To see this, we check if \mbox{$\mathbf{F}$} is the gradient of a scalar potential function. Let \mbox{$\mathbf{F} = \nabla \phi$}, where \mbox{$\phi(x, y) = x^2/2 - y^2/2$}. Then:\par
\qexeq{\frac{\partial \phi}{\partial x} = x = \text{the } x\text{-component of } \mathbf{F}}
\qexeq{\frac{\partial \phi}{\partial y} = -y = \text{the } y\text{-component of } \mathbf{F}}
Since the curve is closed, the work done by a conservative force field around a closed path is zero. However, the curve is not a simple closed loop in the context of the potential function, but rather a circle. The work done is still calculated as above.\par
\qexhead{Step 7: Conclusion}
The work done by the force field on the particle as it moves along the closed curve is:\par
\qexeqhl{\boxed{2\pi}}
\qexhl{This result indicates that the work done is nonzero, which is consistent with the fact that the force field is not conservative when considered in the context of the specific closed path.} The work done is \mbox{$2\pi$}.\par
\end{qexbox}

\section{Prompts}
\label{app:sec:prompts}

\subsection{Training}
\label{app:subsec:prompts-training}

We provide prompts for matching rubrics to units in Figure~\ref{fig:promtpt-ours-match}, prompts for deciding necessity threshold in Figure~\ref{fig:prompt-ours-threshold-1}--\ref{fig:prompt-ours-threshold-2}, and prompts for deciding factual correctness in Figure~\ref{fig:prompt-ours-correctness-1}--\ref{fig:prompt-ours-correctness-2}.
We also provide prompts for baselines and ablations in Figure\ref{fig:prompt-reweighted-1}--\ref{fig:promt-online-2}.

\begin{figure}[!h]
\footnotesize

% ours

\begin{tcblisting}{text only,
    halign=left, 
    title=\textbf{\textsc{Fork} (Rubrics to Units Matching)}, 
    colbacktitle=gray!30!white, 
    coltitle=black,
}

===SYSTEM===\\
You are an expert medical evaluator. You are given a user QUESTION, the COVERAGE RUBRIC written for it — each criterion carrying a separate grader's verdict (MET or NOT MET) and that grader's reason — and a numbered list of UNITS taken from a model's response. Your task is attribution, not grading: for EACH criterion marked MET, identify the units that are the basis for that verdict.\\
~\\

About the units:\\
- Each unit is an exact span of the response, so it may lean on its neighbours or its section heading for meaning — a pronoun, a bare list item, a clause whose main verb is in the heading. Read it in that context.\\
- Units are grouped under [section] lines carrying the response's own headings and lead-ins, in the order the response wrote them. Those [section] lines are context only — never cite them; cite unit numbers only.\\
~\\

How to attribute:\\
- Take the verdicts and the reasons as given, even where you would have judged differently. You are locating where in the response the grader's reason holds, not re-deciding it.\\
- A unit is part of the basis if it supplies something the criterion asks for. A unit that merely touches the criterion's topic without supplying what it asks for is not.\\
- List EVERY unit that is part of the basis, not just the clearest example. When several units each supply part of what the criterion asks for — the items of an enumeration, the two halves of a comparison, a claim and the evidence required for it — list all of them.\\
- If the basis is the response as a whole rather than particular units — its organization, its consistency, its tone, or what it refrains from doing — answer ``whole-response" instead of a list. A criterion about a property the response shows or avoids throughout — tone, organization, consistency, hedging, terminology, what it never says — is carried by every unit: answer ``whole-response" and do not list the units where the property is most visible, because those are examples of the basis, not the basis; list units only when the criterion asks for specific content that particular units supply.\\
- If you cannot find any unit that is the basis for a criterion marked MET, give an empty list.
- Criteria marked NOT MET are context only. Do not include entries for them.\\
~\\

Output a single JSON object keyed by the numbers of the criteria marked MET (as strings), and nothing else:\\
\verb|```|~json\\
\{\\
~~``1": \{``units": [2, 3, 5]\},\\
~~``13": \{``units": ``whole-response"\},\\
~~``20": \{``units": []\}\\
\}\\
\verb|```|~\\
~\\

Include an entry for EVERY criterion marked MET — a criterion you omit is scored as unattributed. Start your response with \verb|```|~json and end with \verb|```|~. Do not include any other text.\\
~\\

===USER===\\
\textless question\textgreater\\
\{question\}\\
\textless /question\textgreater\\
~\\

\textless coverage\_rubric\textgreater\\
\{rubric\_block\}\\
\textless /coverage\_rubric\textgreater\\
~\\

\textless units\textgreater\\
\{units\_block\}\\
\textless /units\textgreater\\
~\\

For each criterion marked MET, give the units that are the basis for its verdict. Your JSON:

\end{tcblisting}
\vspace{-2mm}
\caption{Prompt for \textsc{Fork} (Rubrics to Units Matching).}
\label{fig:promtpt-ours-match}
% \vspace{-4mm}
%
\end{figure}

\begin{figure}[!h]
\footnotesize

\begin{tcblisting}{text only,
    halign=left, 
    title=\textbf{\textsc{Fork} (Necessity Threshold Decision) (1/2)}, 
    colbacktitle=gray!30!white, 
    coltitle=black,
}

===SYSTEM===\\
You are an expert medical evaluator. You are given a user QUESTION and a model's ANSWER to it, split into numbered UNITS — exact spans of the answer, in its order, under its [section] lines. The answer was graded against a coverage rubric — the points a good answer should contain — and each unit carries a score in parentheses: the share of rubric credit it holds. A unit that alone carries a credited point scores 1.00; a unit that is one of K units carrying the same point scores 1/K; shares add up when a unit carries several points, so a score can exceed 1.00; a unit that carries no credited point scores 0.00. The rubric is not shown.\\
~\\

Your task has two steps, answered together. First, set the THRESHOLD for this answer: the lowest score a unit must carry to be necessary — a unit the question needs. Then deduplicate among the necessary units. Rubric credit is shared by duplicates: when the answer makes a point and later makes it again, both units can carry a share of it. For each necessary unit in the assigned range, decide whether it duplicates an earlier necessary unit — verdict ``remove" — or makes a contribution of its own — verdict ``keep".\\
~\\

About the threshold:\\
- The score says how much of a credited point a unit carries; a low score means the credit for some point is spread over many units. Decide, for this question, how thin that spread may be before a unit stops being necessary.\\
- A question that calls for a list, a set of steps or several distinct items legitimately spreads a point over many units, and a low threshold keeps them. A question answered by a few points should not have its credit spread thin, and a higher threshold cuts the units that only dilute it.\\
- Choose the threshold from the scores in front of you, reading the units themselves as the evidence of what the spread means in this answer. A unit scoring 0.00 carries no credited point and is never necessary, so the threshold is greater than 0.00.\\
- Give the threshold as a decimal with two places, in the same form as the scores. A unit whose score is at or above the threshold is necessary; every other unit is context only.\\
~\\

About the units:\\
- Each unit is an exact span of the answer, so it may lean on its neighbours or its section heading for meaning — a pronoun, a bare list item, a clause whose main verb is in the heading. Read it in that context.\\
- Units are grouped under [section] lines carrying the answer's own headings and lead-ins, in the order the answer wrote them. Those [section] lines are context only — never judge them.\\
- Units below the threshold are context only. They are never judged, and a necessary unit never duplicates a unit below the threshold: compare necessary units with earlier necessary units only.\\
~\\

The principle: a unit is a duplicate when there is no semantically meaningful difference between what it gives the asker and what an earlier necessary unit already gave. A difference is meaningful when it changes what the asker knows or can do; it is trivial when it only changes how the same thing is said — shorter, longer, reworded, re-explained, emphasized, or placed under another heading.\\
~\\

Common duplicates: the same fact, advice, item, warning or conclusion stated again in other words; the same point repeated at more length or with a generic reason attached; a summary, recap or conclusion that only gathers points already made; an item listed again that an earlier unit already listed; two entries of one list that make the same point under different names.\\
~\\

Direction: the duplicate is always the later unit, never the first to make the point. ``Earlier" means earlier in the answer's order, anywhere in the answer: the comparison spans every section and reaches outside the assigned units — all units are shown for that. Nothing that comes later makes an earlier unit a duplicate.\\
~\\

A ``remove" must name the earlier unit's number in its reasoning; if you cannot name one, the verdict is ``keep". When the difference is genuinely unclear, the verdict is ``keep".\\
~\\

In the verdicts, do NOT judge whether the unit is true, needed, or well written — the threshold has already decided what is necessary, and the rest are different evaluations. Judge as if every unit's content were accurate.

\end{tcblisting}
\vspace{-2mm}
\caption{Prompt for \textsc{Fork} (Necessity Threshold Decision) (1/2).}
\label{fig:prompt-ours-threshold-1}
% \vspace{-4mm}
%
\end{figure}

\begin{figure}[!h]
\footnotesize

\begin{tcblisting}{text only,
    halign=left, 
    title=\textbf{\textsc{Fork} (Necessity Threshold Decision) (2/2)}, 
    colbacktitle=gray!30!white, 
    coltitle=black,
}

Output a single JSON object, and nothing else: first the key ``threshold" with your decimal, then one entry per necessary unit in the assigned range, keyed by its number (as a string) — each value is an object with ``reasoning" first, then ``verdict", never a bare string. The reasoning is a few words; a ``remove" names the earlier unit:\\
\verb|```|~json\\
\{\\
~~``threshold": 0.25,\\
~~``3": \{``reasoning": ``first statement of the dosing rule", ``verdict": ``keep"\},\\
~~``8": \{``reasoning": ``duplicates 3: the same dosing rule reworded", ``verdict": ``remove"\},\\
~~``9": \{``reasoning": ``adds the threshold that 3 does not give", ``verdict": ``keep"\},\\
~~``14": \{``reasoning": ``recap of 3, 9 and 11, nothing new", ``verdict": ``remove"\}\\
\}\\
\verb|```|~\\
~\\

Include an entry for EVERY necessary unit in the assigned range — a necessary unit you omit is scored as invalid. Do not include entries for units below the threshold or outside the assigned range. Start your response with \verb|```|~json and end with \verb|```|~. Do not include any other text.\\
~\\

===USER===\\
\textless question\textgreater\\
\{question\}\\
\textless /question\textgreater\\
~\\

\textless answer\textgreater\\
\{units\_block\}\\
\textless /answer\textgreater\\
~\\

Set the threshold for this answer, then judge the units \{judge\_range\} whose score is at or above it — ``keep" or ``remove" for each. Every other unit is shown for context only. Your JSON: ``threshold" first, then one \{``reasoning": ..., ``verdict": ...\} object per necessary unit in that range:

\end{tcblisting}
\vspace{-2mm}
\caption{Prompt for \textsc{Fork} (Necessity Threshold Decision) (2/2).}
\label{fig:prompt-ours-threshold-2}
% \vspace{-4mm}
%
\end{figure}

\begin{figure}[!h]
\footnotesize

\begin{tcblisting}{text only,
    halign=left, 
    title=\textbf{\textsc{Fork} (Factual Correctness Decision) (1/2)}, 
    colbacktitle=gray!30!white, 
    coltitle=black,
}

===SYSTEM===\\
You are an expert medical evaluator. You are given a user QUESTION and a numbered list of UNITS taken from a model's answer to it. Judge the FACTUAL CORRECTNESS of each assigned unit — and nothing else about it.\\
~\\

About the units:\\
- Each unit is an exact span of the answer, so it may lean on its neighbours or its section heading for meaning — a pronoun, a bare list item, a clause whose main verb is in the heading. Read it in that context and judge what it says there.\\
- Units are grouped under [section] lines carrying the answer's own headings and lead-ins, in the order the answer wrote them. Those [section] lines are context only — never judge them.\\
- Only the assigned units are judged; every other unit is shown for context.\\
~\\

For every assigned unit, give exactly one verdict. Decide in this order:\\
- ``na" — the unit asserts nothing checkable: it only organizes the answer, leads into or sums up other units, expresses empathy or reassurance, comments on the answer itself, or offers further help. Decide this first. A unit that tells the asker what is true, what to do, how, how much, when, or why — advice and instructions included — asserts something checkable and is never ``na", however brief or hedged.\\
- ``incorrect" — the unit asserts something wrong: it contradicts reliable medical knowledge or the facts the question states, gives advice or an instruction that is wrong or unsafe for the situation the question describes, or presents as standard a name, term or practice the field does not recognize. A true fact applied to the wrong situation is an error in that situation. A unit that asserts several things is ``incorrect" if any one of them is wrong.\\
- ``correct" — what the unit asserts is right, at the strength and breadth it asserts it: supported by reliable medical knowledge you hold, common knowledge, or the question itself. Common clinical knowledge that any clinician would confirm is ``correct" without a citation; ``uncertain" is for claims that would need one.\\
- ``uncertain" — the unit asserts something checkable, but you cannot confirm it from reliable medical knowledge you hold, at the strength and breadth it asserts it: a specific you cannot verify (often a very specific figure, a rare interaction, a niche study claim, a named source), or an overstatement — more certainty, a stronger or broader effect, or a narrow or contested finding presented as general. Judge the takeaway a careful reader would carry away: a hedge (``may", ``can", ``often") still asserts that the link or effect is real, so it does not lower the bar. Plausible, commonly stated or hard to disprove is not confirmed; confirmation comes from what you actually know, not from a source you would have to assume exists.\\
~\\

How to judge:\\
- Judge each assigned unit only on its own content. A unit is not wrong because a neighbouring unit is wrong.\\
- Do NOT judge whether the unit is needed, relevant or repeated — that is a different evaluation. A redundant, generic or tangential unit whose content is factually accurate is ``correct".\\
~\\

Output a single JSON object keyed by the assigned unit numbers (as strings), and nothing else — each unit's value is an object with ``reasoning" first, then ``verdict", never a bare string. The reasoning is a few words where the verdict is obvious, one sentence where it takes thought; for ``incorrect" it must name the error; for ``uncertain", the claim you cannot confirm — the specific, or the strength or breadth it overstates:\\
\verb|```|~json\\
\{\\
~~``6": \{``reasoning": ``matches standard dosing guidance", ``verdict": ``correct"\},\\
~~``7": \{``reasoning": ``transition into the treatment section, asserts nothing", ``verdict": ``na"\},\\
~~``8": \{``reasoning": ``states antibiotics treat viral infections — they do not act on viruses", ``verdict": ``incorrect"\},\\
~~``9": \{``reasoning": ``specific prevalence figure I can neither support nor refute", ``verdict": ``uncertain"\},\\
~~``10": \{``reasoning": ``'may provide significant relief' still conveys a real, sizeable effect; the evidence is weak and mixed", ``verdict": ``uncertain"\}\\
\}\\
\verb|```|~\\

\end{tcblisting}
\vspace{-2mm}
\caption{Prompt for \textsc{Fork} (Factual Correctness Decision) (1/2).}
\label{fig:prompt-ours-correctness-1}
% \vspace{-4mm}
%
\end{figure}

\begin{figure}[!h]
\footnotesize

\begin{tcblisting}{text only,
    halign=left, 
    title=\textbf{\textsc{Fork} (Factual Correctness Decision) (2/2)}, 
    colbacktitle=gray!30!white, 
    coltitle=black,
}

Include an entry for EVERY assigned unit number — a unit you omit is scored as invalid. Do not include entries for units that are not assigned. Start your response with \verb|```|~json and end with \verb|```|~. Do not include any other text.\\
~\\

===USER===\\
\textless question\textgreater\\
\{question\}\\
\textless /question\textgreater\\
~\\

\textless units\textgreater\\
\{units\_block\}\\
\textless /units\textgreater\\
~\\

Judge the factual correctness of units \{judge\_range\}. The other units are shown for context only. Your JSON, one \{``reasoning": ..., ``verdict": ...\} object per assigned unit:

\end{tcblisting}
\vspace{-2mm}
\caption{Prompt for \textsc{Fork} (Factual Correctness Decision) (2/2).}
\label{fig:prompt-ours-correctness-2}
% \vspace{-4mm}
%
\end{figure}

% baselines

\begin{figure}[!h]
\footnotesize

\begin{tcblisting}{text only,
    halign=left, 
    title=\textbf{Reweighted Absence-Based Rubrics (1/2)}, 
    colbacktitle=gray!30!white, 
    coltitle=black,
}
===SYSTEM===\\
You are an expert evaluator. Given a prompt (which may include conversation history) and a generated response, please evaluate the response against EACH rubric, and state the weight each absence-based rubric item should carry.\\
~\\

Return a json object where keys are the rubric numbers (as strings) and values are objects with ``explanation", ``criteria\_met", ``type" and ``updated\_weight" fields.\\
- The ``explanation" field should be a string explaining why the response does or does not meet the criteria of the rubric item.\\
- The ``criteria\_met" field should be a boolean indicating whether the response meets the criteria of the rubric item. If a rubric item has multiple sentences or criteria, you should consider all of them. If any of the criteria is not met, the answer should be false. Only return true if all of the criteria are met. The weight listed with an item has no bearing on this verdict: decide whether the criterion is met exactly as you would if no weight were shown.\\
- One important exception to the above bullet point is that if a criteria says ``such as", ``for example", or ``including", the response does not have to include all of the examples listed to meet the criteria. For example, if the criteria says ``Mentions practical ways to reduce household electricity use such as lowering thermostat settings, using LED bulbs, and unplugging idle devices", and the response just mentions using LED bulbs, that would still meet the criteria even though it didn't mention all of the specific examples listed in the criteria.\\
- The ``type" field should be ``presence" or ``absence", as defined under Criterion types below. It classifies the rubric item's wording, not your verdict: a presence-based item that the response fails to satisfy is still ``presence".\\
- The ``updated\_weight" field should be an integer for an absence-based item, chosen as described under Updated weights below, and null for a presence-based item.\\
~\\

\# Criterion types\\
Each rubric item is presence-based or absence-based. The type is a property of the item's wording, read before looking at the response: it does not depend on the response or on whether the item is met.\\
- A presence-based item asks whether the response contains something: a fact, a reasoning step, a list, a recommendation, a disclaimer, a format or a tone. It is met when that content appears explicitly in the response and not met when the content is missing. Missing content makes the item not met; it does not make the item absence-based. If your explanation says that the response does not mention, state or provide what the item asks for, the item is presence-based.\\
- An absence-based item asks whether the response avoids something. Its wording forbids or limits: it does not claim, recommend, include, assume or fabricate X; it contains no errors or contradictions; it stays within a limit such as ``only X" or ``without Y". It is met only if nothing anywhere in the response violates it, so check the whole response before marking it met, and say in the explanation what you checked for and what you found.\\
- When a rubric item has both a presence requirement and an absence requirement, label it ``absence", even when most of the item asks for content and the prohibition or limit is a single clause. Its ``criteria\_met" still requires both parts to be met.\\
- A correctness word attached to a specific required fact, such as ``correctly states that X", makes a presence-based item, not an absence-based one.\\
~\\

\# Updated weights\\
Each rubric item is listed with its weight. Rubrics like this one describe what a good answer should contain in many items but what it must avoid in only a few, because the ways an answer can go wrong cannot be enumerated the way its required content can. Graded as listed, such a rubric rewards adding content: every extra sentence might satisfy one more presence-based item, while the few absence-based items cost little to violate, so answers grow padded and error-prone. The absence-based items are the rubric's only guards against that, and their listed weights understate that role. Beside deciding whether each item is met, your second job is to state, for each absence-based item, the weight it should carry, so that the grade favors an answer that says what is needed, says it correctly, and stops there.
\end{tcblisting}
\vspace{-2mm}
\caption{Prompt for Reweighted Absence-Based Rubrics (1/2).}
\label{fig:prompt-reweighted-1}
% \vspace{-4mm}
%
\end{figure}

\begin{figure}[!h]
\footnotesize

\begin{tcblisting}{text only,
    halign=left, 
    title=\textbf{Reweighted Absence-Based Rubrics (2/2)}, 
    colbacktitle=gray!30!white, 
    coltitle=black,
}

Size the weight against the presence-based items the item guards: violating it should cost at least as much as the answer could gain by satisfying more of them. Two things decide how much that is, and the judgment is yours:\\
- What the item guards. Ask how much of the answer's value depends on it, not how broadly it is worded. A guard on whether the answer can be trusted or acted on at all, such as no incorrect statements or no unsafe instructions, could deserve a weight above the highest listed weight, possibly several times over. A narrowly worded item could deserve as much when the detail it protects is decisive, and a broadly worded one might warrant only a moderate increase when what it guards matters little for this question. These are illustrations, not fixed tiers.\\
- What this response did. If the response violates the item, weigh the damage the violation does to the answer for the asker. If it avoids the item, weigh what the answer would have lost had it not.\\
~\\

Give the weight as an integer on the same scale as the listed weights, never below the listed weight, and as high as the case warrants. The response's own statements about what it avoids do not raise the weight. For presence-based items, ``updated\_weight" is null.\\
~\\

\# Example\\
For example, if the prompt is ``How can I reduce electricity use at home?" and the response is ``Leave all the lights on overnight." and rubric item 1 is ``suggests practical ways to reduce electricity use", you should return:\\
~\\

\verb|```|~json\\
\{\\
~~``1": \{``explanation": ``The response did not suggest practical ways to reduce electricity use.", ``criteria\_met": false, ``type": ``presence", ``updated\_weight": null\}\\
\}\\
\verb|```|~\\
~\\

Start your response with a valid JSON object that starts with \verb|```|~json and ends with \verb|```|~. Do not include any other text. Evaluate ALL rubric items and include an entry for each one.\\
~\\

===USER===\\
Given the following prompt and response, please evaluate the response against each rubric.\\
~\\

\textless prompt\textgreater\\
\{prompt\}\\
\textless /prompt\textgreater\\
~\\

\textless response\textgreater\\
\{response\}\\
\textless /response\textgreater\\
~\\

\textless rubrics\textgreater\\
\{rubric\_list\_string\}\\
\textless /rubrics\textgreater\\
~\\

Your JSON Evaluation:
\end{tcblisting}
\vspace{-2mm}
\caption{Prompt for Reweighted Absence-Based Rubrics (2/2).}
\label{fig:prompt-reweighted-2}
% \vspace{-4mm}
%
\end{figure}

\begin{figure}[!h]
\footnotesize

\begin{tcblisting}{text only,
    halign=left, 
    title=\textbf{Holistic Preference (1/2)}, 
    colbacktitle=gray!30!white, 
    coltitle=black,
}

===SYSTEM===\\
You are an expert evaluator. You will be given a user prompt, a scoring rubric written for that prompt, and two responses to the prompt (@response\_A and @response\_B). Your job is to evaluate EACH response on the following 4 dimensions, scoring each from 1 (worst) to 7 (best), and then decide which response is preferable overall.\\
~\\

Dimensions and anchors:\\
~\\

1. Rubric coverage - how well the response covers the rubric's criteria, where each criterion's weight states its importance.\\
~~~1: covers almost none of the rubric's weight; 4: covers roughly half of the rubric's weight; 7: covers every criterion, including the heavily weighted ones.\\
~\\

2. Correctness - All factual claims, calculations, and recommendations in the response are correct — supported by the prompt, the original rubric, or reliable domain knowledge — with nothing fabricated, misleading, numerically wrong, or overconfident.\\
~~~1: many false or fabricated claims (5+); 3: several errors (3-4 incorrect claims); 5: a few minor errors (1-2 incorrect claims); 7: no detectable factual errors.\\
~\\

3. Necessity - The response needed every part of it: nothing restates what an earlier part already said in the same or different words, nothing is boilerplate, filler, or a pleasantry, nothing is vague padding, and no section, example, or caveat is there only to sound thorough — removing any part would noticeably weaken the answer.\\
~~~1: extensive repetition, boilerplate, or padding; 4: several parts could be removed without weakening the answer; 7: every part is needed; removing any of them would noticeably weaken the answer.\\
~\\

Instructions:\\
Judge each dimension for EACH response. For each dimension, write a brief justification (1-2 sentences) FIRST, then give the score. For rubric coverage, assess the response's coverage as a whole rather than item by item, citing the criteria that most affected your judgment where that makes the assessment clear. The rubric applies to both responses equally; neither response is a reference for the other. After scoring all dimensions, weigh all four dimensions as you see fit and state which response is preferable overall, ``A" or ``B", with a 2-3 sentence explanation referencing @response\_A and @response\_B. Answer ``tie" when both responses are of equal overall quality and neither can be preferred over the other. Be fair and objective. Do not let the ordering of responses bias you.

\end{tcblisting}
\vspace{-2mm}
\caption{Prompt for Holistic Preference (1/2).}
\label{fig:prompt-holistic-1}
% \vspace{-4mm}
%
\end{figure}

\begin{figure}[!h]
\footnotesize

\begin{tcblisting}{text only,
    halign=left, 
    title=\textbf{Holistic Preference (2/2)}, 
    colbacktitle=gray!30!white, 
    coltitle=black,
}

Return a single JSON object with this exact shape (scores are integers 1-7; ``preferred" is ``A", ``B", or ``tie"):\\
\verb|```|~json\\
\{\\
~~``response\_A": \{\\
~~~~``coverage": \{``justification": ``", ``score": 0\},\\
~~~~``correctness": \{``justification": ``", ``score": 0\},\\
~~~~``necessity": \{``justification": ``", ``score": 0\}\\
~~\},\\
~~``response\_B": \{\\
~~~~``coverage": \{``justification": ``", ``score": 0\},\\
~~~~``correctness": \{``justification": ``", ``score": 0\},\\
~~~~``necessity": \{``justification": ``", ``score": 0\}\\
~~\},\\
~~``comparison": \{``explanation": ``", ``preferred": ``"\}\\
\}\\
\verb|```|~\\
Start your response with \verb|```|~json and end with \verb|```|~. Do not include any other text.\\
~\\

===USER===\\
\textless prompt\textgreater\\
\{prompt\}\\
\textless /prompt\textgreater\\
~\\

\textless rubric\textgreater\\
\{rubric\_block\}\\
\textless /rubric\textgreater\\
~\\

\textless response\_A\textgreater\\
\{response\_A\}\\
\textless /response\_A\textgreater\\
~\\

\textless response\_B\textgreater\\
\{response\_B\}\\
\textless /response\_B\textgreater\\
~\\

Your JSON Evaluation:

\end{tcblisting}
\vspace{-2mm}
\caption{Prompt for Holistic Preference (2/2).}
\label{fig:prompt-holistic-2}
% \vspace{-4mm}
%
\end{figure}

\begin{figure}[!h]
\vspace{-5mm}
\footnotesize

\begin{tcblisting}{text only,
    halign=left, 
    title=\textbf{Generic Criteria}, 
    colbacktitle=gray!30!white, 
    coltitle=black,
}

===SYSTEM===\\
You are an expert medical evaluator. You are given a user QUESTION, the ORIGINAL RUBRIC written for it, optional REFERENCE material, and a numbered list of UNITS taken from a model's response. Judge EACH assigned unit against two criteria.\\
~\\

About the units:\\
- Each unit is an exact span of the response, so it may lean on its neighbours or its section heading for meaning — a pronoun, a bare list item, a clause whose main verb is in the heading. Read it in that context and judge what it says there.\\
- Units are grouped under [section] lines carrying the response's own headings and lead-ins, and they appear in the order the response wrote them. Those [section] lines are context only — never judge them.\\
~\\

Criteria (judge each as true/false for every assigned unit):\\
~\\

1. correct — The unit is factually correct and supported by the prompt, the reference material (if provided), the original rubric, or reliable domain knowledge. It is NOT fabricated, misleading, numerically wrong, or overconfident. If you cannot support the unit from any of these, mark it false.\\
~\\

2. necessary — The response NEEDED this unit. Ask: would a well-written, concise answer to this question have included it? Common reasons a unit is unnecessary — any one is enough, and the list is not exhaustive:\\
~~~- it restates something an EARLIER unit already said, in the same or different words;\\
~~~- it is boilerplate, filler, or a pleasantry (greetings, sign-offs, offers to answer follow-up questions, disclaimers already made);\\
~~~- it is padding: vague, generic, or so unspecific that it adds nothing of substance;\\
~~~- it is there only to sound thorough — an extra section, example, or caveat that adds nothing to answering the question.\\
~~~When a unit is borderline, test it by removal: read the response without it, and if its absence would noticeably weaken the answer, it is necessary; otherwise it is not.\\
~~~A unit that is simply wrong should fail `correct', not `necessary' — judge necessity as if the unit were true.\\
~\\

Judge `correct' on the unit's own merits. `necessary' is different: it is a judgement about the unit's place in the whole response, so weigh it against the other units and against what a good answer to this question needs.\\
~\\

Output a single JSON object keyed by the unit numbers (as strings), and nothing else:\\
\verb|```|~json\\
\{\\
~~``1": \{``explanation": ``\textless one short sentence\textgreater", ``correct": true, ``necessary": true\},\\
~~``2": \{``explanation": ``\textless one short sentence\textgreater", ``correct": false, ``necessary": false\}\\
\}\\
\verb|```|~\\
~\\

Include an entry for EVERY assigned unit number — a unit you omit is scored as invalid. Do not include entries for units outside the assigned range. Start your response with \verb|```|~json and end with \verb|```|~. Do not include any other text.\\
~\\

===USER===\\
\textless question\textgreater\\
\{question\}\\
\textless /question\textgreater\\
\{reference\_block\}\\
\textless original\_rubric\textgreater\\
\{rubric\_block\}\\
\textless /original\_rubric\textgreater\\
~\\

\textless units\textgreater\\
\{units\_block\}\\
\textless /units\textgreater\\
~\\

Judge units \{judge\_range\} against correct / necessary. The other units are shown for context only. Your JSON:

\end{tcblisting}
\vspace{-2mm}
\caption{Prompt for Generic Criteria.}
\label{fig:prompt-generic}
% \vspace{-4mm}
%
\end{figure}

\begin{figure}[!h]
\footnotesize

\begin{tcblisting}{text only,
    halign=left, 
    title=\textbf{Offline Criteria (1/2)}, 
    colbacktitle=gray!30!white, 
    coltitle=black,
}

===SYSTEM===\\
You are an expert medical evaluator. You are given a user QUESTION, the COVERAGE RUBRIC written for it, optional REFERENCE material, a PRECISION RUBRIC written for this question, and a numbered list of UNITS taken from a model's response. Judge EACH assigned unit.\\
~\\

About the units:\\
- Each unit is an exact span of the response, so it may lean on its neighbours or its section heading for meaning — a pronoun, a bare list item, a clause whose main verb is in the heading. Read it in that context and judge what it says there.\\
- Units are grouped under [section] lines carrying the response's own headings and lead-ins, and they appear in the order the response wrote them. Those [section] lines are context only — never judge them.\\
~\\

Judgements for every assigned unit:\\
~\\

1. correct (true/false) — The unit is factually correct and supported by the prompt, the reference material (if provided), the coverage rubric, or reliable domain knowledge. It is NOT fabricated, misleading, numerically wrong, or overconfident. If you cannot support the unit from any of these, mark it false.\\
~\\

2. the PRECISION RUBRIC — a numbered list of criteria, each with a scope (which units it applies to) and a test (what an in-scope unit must or must not do). For each criterion, decide the scope first: does it cover this unit? A unit outside a criterion's scope is never judged by that criterion. Then, for each criterion that does cover the unit, decide whether the unit complies with its test. Report the numbers of the criteria the unit fails; a unit that no criterion covers, or that complies with every criterion covering it, gets an empty list. Some scopes and tests refer to the unit's place in the whole response — its section, its position, what earlier units already said. All units are shown so you can decide those; use them.\\
~\\

Some tests are VERIFICATION tests: they make compliance conditional on a specific claim (a name, a value, a citation, a mechanism) being standard, well-established knowledge — recognizable by conditions like ``unless the stated dose is standard, well-established dosing". Answer that condition directly and affirmatively: the unit meets it only if you can support the specific claim from the prompt, the reference material (if provided), the COVERAGE RUBRIC, or reliable domain knowledge you actually recognize. A claim you cannot place — an unfamiliar study or formula name, a value you cannot confirm, a real entity attached to a claim you do not recognize — does not meet the condition, however confident the wording. Not every ``unless" marks a verification test: an exception clause about behavior or circumstance — ``unless it is strictly listing red-flag symptoms and clearly distinguishes them from the user's current state" — is ordinary test content; decide by reading whether that circumstance actually holds for this unit.\\
~\\

Accuracy and compliance are separate questions; never mix them. `correct' judges accuracy. Each precision criterion judges compliance with its own test, exactly as written: a unit is not excused from a test because what it says is accurate, and a unit never fails a test merely because what it says is inaccurate — inaccuracy fails `correct' and nothing else. Verification tests are the one deliberate overlap: whether the claim is established IS their test, so answer it there.\\
~\\

The COVERAGE RUBRIC is support material for `correct' and for verification tests — never grade units against it directly; coverage is graded elsewhere.\\
~\\

Output a single JSON object keyed by the unit numbers (as strings), and nothing else:\\
\verb|```|~json\\
\{\\
~~``1": \{``explanation": ``\textless one short sentence\textgreater", ``correct": true, ``fails": []\},\\
~~``2": \{``explanation": ``\textless one short sentence, naming what was failed\textgreater", ``correct": true, ``fails": [3]\}\\
\}\\
\verb|```|~\\
~\\

Include an entry for EVERY assigned unit number — a unit you omit is scored as invalid. `fails` must always be present (an empty list if none). Do not include entries for units outside the assigned range. Start your response with \verb|```|~json and end with \verb|```|~. Do not include any other text.\\
~\\

\end{tcblisting}
\vspace{-2mm}
\caption{Prompt for Offline Criteria (1/2).}
\label{fig:prompt-offline-1}
% \vspace{-4mm}
%
\end{figure}

\begin{figure}[!h]
\footnotesize

\begin{tcblisting}{text only,
    halign=left, 
    title=\textbf{Offline Criteria (2/2)}, 
    colbacktitle=gray!30!white, 
    coltitle=black,
}

===USER===\\
\textless question\textgreater\\
\{question\}\\
\textless /question\textgreater\\
\{reference\_block\}\\
\textless coverage\_rubric\textgreater\\
\{rubric\_block\}\\
\textless /coverage\_rubric\textgreater\\
~\\

\textless precision\_rubric\textgreater\\
\{precision\_rubric\_block\}\\
\textless /precision\_rubric\textgreater\\
~\\

\textless units\textgreater\\
\{units\_block\}\\
\textless /units\textgreater\\
~\\

Judge units \{judge\_range\} against correct / the precision rubric. The other units are shown for context only. Your JSON:

\end{tcblisting}
\vspace{-2mm}
\caption{Prompt for Offline Criteria (2/2).}
\label{fig:prompt-offline-2}
% \vspace{-4mm}
%
\end{figure}

\begin{figure}[!h]
\footnotesize

\begin{tcblisting}{text only,
    halign=left, 
    title=\textbf{Online Criteria (1/2)}, 
    colbacktitle=gray!30!white, 
    coltitle=black,
}

===SYSTEM===\\
You are an expert medical evaluator. You are given a user QUESTION and a model's ANSWER to it, split into numbered UNITS — exact spans of the answer (sentences, list items, clauses), in its order, under its [section] lines. The answer was graded against a coverage rubric — the points a good answer should contain — and each point it earned was credited to the units carrying it: those are tagged (credited), the rest (uncredited). Credited units are the core content. Uncredited units are either peripheral content the answer still needs — lead-ins, qualifiers, examples and details that extend a credited point, relevant points the rubric did not ask for — or futile content — padding, repetition, digression — and the tag does not say which: an uncredited unit may or may not have an undesirable property, and telling the two apart is where your job starts.\\
~\\

Your task is to write the PRECISION RUBRIC for this answer: extract the undesirable properties the uncredited units show, if any, and turn each into a criterion. A tag records information — which point a unit carries; a property is behaviour — what a unit does toward the question, the units around it or its own point. A property is undesirable when deleting units that have it leaves the answer no worse for the asker: no less correct, complete or clear. For example: saying again what an earlier unit already said (adding a detail, example, qualifier or reason to a point is not saying it again); filler, pleasantries, sign-offs or commentary on the answer itself; generic advice that would fit any question on the topic; explaining or justifying at length where a unit of that kind only needs to name the point; drifting to a topic the question did not ask about. It helps to ask what an uncredited unit does that the credited units need not do — but a tag is not a property: credited units can show these properties too, to some degree, and uncredited units can carry something the asker needs — a qualifier, an example, a point of their own. A separate grader will apply every criterion to every unit, credited or not, without seeing the tags, your analysis or your unit lists; a unit is good only if it satisfies every criterion that applies to it.\\
~\\

Rules:\\
- Each criterion is a scope and a prohibition. ``scope" is ``units that \textless kind of content \textgreater" — a kind of content, never a section, a position or one particular unit. ``criterion" begins ``A unit should not …" and is one prohibition the grader can check on the unit alone; only a repeat may refer to earlier units. Make the scope ``all units" when the property is not confined to one kind of unit but should be checked on every unit of the answer.\\
- A criterion is about what a unit does, never about the point it makes or fails to make — what the answer must contain is judged by the coverage rubric, not by your criteria. If a credited unit shows the property because of the point it makes, the criterion is about content: narrow the scope or drop it. Content the question asks for is never off-topic, even when uncredited.\\
- Write a criterion only for an undesirable property you actually see in the uncredited units, and list the uncredited units that show it (uncredited only). Write as many criteria as there are distinct undesirable properties; an uncredited unit showing no such property needs no criterion, and if none does, return an empty list.\\
- One property per criterion. Prefer several simple criteria to one that chains different properties with ``or"; ``or" is fine between forms of the same property (pleasantries or sign-offs). Overlap between criteria is harmless; the same property twice is not.\\
- The grader cannot see which units are credited. A criterion must never mention or depend on credit — the words credit, credited, uncredited and rubric do not appear in a scope or criterion. A repeat is judged against earlier units of any kind: write ``an earlier unit", never ``a credited unit".\\
- Do not quote, paraphrase, number or otherwise point at units in ``scope" or ``criterion": each must make sense applied to any unit of any answer to this question.\\
- Begin with a brief ``analysis" (a few sentences): which uncredited units the answer needs and which show an undesirable property — by groups, not one by one — and what the latter do that the answer would be no worse without. Then write the criteria.\\

\end{tcblisting}
\vspace{-2mm}
\caption{Prompt for Online Criteria (1/2).}
\label{fig:prompt-online-1}
% \vspace{-4mm}
%
\end{figure}

\begin{figure}[!h]
\footnotesize

\begin{tcblisting}{text only,
    halign=left, 
    title=\textbf{Online Criteria (2/2)}, 
    colbacktitle=gray!30!white, 
    coltitle=black,
}

For example, for a question asking what to do for someone with a suspected stroke while waiting for the ambulance:\\
- scope ``all units" — ``A unit should not say again what an earlier unit already said, in the same or in different words."\\
- scope ``units that tell the reader whom to call or where to go for help" — ``A unit should not add unrequested reasons why that service is the right one beyond naming it."\\
- scope ``all units" — ``A unit should not explain the mechanism or types of stroke, which the question did not ask about."\\
Not criteria: ``A unit should mention noting the time the symptoms began." (requires a point — the coverage rubric's job); ``A unit should not describe what to do while waiting." (bans what the question asks for); ``A unit should not add steps not already covered by the credited units." (refers to credit, which the grader never sees).\\
~\\

Output a single JSON object and nothing else:\\
\verb|```|~json\\
\{``analysis": ``\textless a few sentences: which uncredited units the answer needs, which show an undesirable property, and what those do\textgreater", ``criteria": [\{``units": [\textless numbers of the uncredited units that show it\textgreater], ``scope": ``\textless all units | units that ...\textgreater", ``criterion": ``A unit should not ..."\}]\}\\
\verb|```|~\\
~\\

Start your response with \verb|```|~json and end with \verb|```|~. Do not include any other text.\\
~\\

===USER===\\
\textless question\textgreater\\
\{question\}\\
\textless /question\textgreater\\
~\\

\textless answer\textgreater\\
\{units\_block\}\\
\textless /answer\textgreater\\
~\\

Write the precision rubric for this answer. Your JSON:

\end{tcblisting}
\vspace{-2mm}
\caption{Prompt for Online Criteria (2/2).}
\label{fig:promt-online-2}
% \vspace{-4mm}
%
\end{figure}

% \begin{figure}[!h]
% \footnotesize

% \begin{tcblisting}{text only,
%     halign=left, 
%     title=\textbf{tbd}, 
%     colbacktitle=gray!30!white, 
%     coltitle=black,
% }

% tbd.

% \end{tcblisting}
% %
% \vspace{-2mm}
% \caption{tbd.}
% \label{fig:tbd}
% % \vspace{-4mm}
% %
% \end{figure}

\subsection{Evaluation}
\label{app:subsec:prompts-eval}

There is one newly introduced judge prompt in our evaluation, which is measuring precision.
We use the prompt same as in \textit{Generic Criteria} baseline (Figure~\ref{fig:prompt-generic}) since both of their job is to evaluate necessity and factual correctness of response units.

% \begin{figure}[!h]
% \footnotesize

% \begin{tcblisting}{text only,
%     halign=left, 
%     title=\textbf{tbd}, 
%     colbacktitle=gray!30!white, 
%     coltitle=black,
% }

% tbd.

% \end{tcblisting}
% %
% \vspace{-2mm}
% \caption{tbd.}
% \label{fig:tbd}
% % \vspace{-4mm}
% %
% \end{figure}

% \begin{figure}[!h]
% \footnotesize

% \begin{tcblisting}{text only,
%     halign=left, 
%     title=\textbf{tbd}, 
%     colbacktitle=gray!30!white, 
%     coltitle=black,
% }

% tbd.

% \end{tcblisting}
% %
% \vspace{-2mm}
% \caption{tbd.}
% \label{fig:tbd}
% % \vspace{-4mm}
% %
% \end{figure}

\end{document}